\documentclass{article}
\usepackage{iclr2027_conference,times}

\usepackage{amsmath,amsfonts,bm}

\def\eqref#1{equation~\ref{#1}}

\def\1{\bm{1}}

\DeclareMathAlphabet{\mathsfit}{\encodingdefault}{\sfdefault}{m}{sl}
\SetMathAlphabet{\mathsfit}{bold}{\encodingdefault}{\sfdefault}{bx}{n}

\usepackage[utf8]{inputenc}
\usepackage[T1]{fontenc}
\usepackage{hyperref}
\usepackage{url}
\usepackage{booktabs}
\usepackage{multirow}
\usepackage{amsfonts}
\usepackage{amsmath}
\usepackage{amssymb}
\usepackage{bbm}
\usepackage{nicefrac}
\usepackage{microtype}
\usepackage{xcolor}
\usepackage{colortbl}
\definecolor{refblue}{RGB}{35,95,185}
\hypersetup{colorlinks=true, citecolor=refblue, linkcolor=refblue, urlcolor=refblue}
\definecolor{emoeblue}{RGB}{31,78,140}
\definecolor{emoeblueborder}{RGB}{18,47,92}
\definecolor{gooddark}{RGB}{0,125,55}
\definecolor{baddark}{RGB}{190,30,30}
\definecolor{mdlmred}{RGB}{217,140,134}
\definecolor{mdlmredborder}{RGB}{168,67,58}
\definecolor{vaddblue}{RGB}{143,181,220}
\definecolor{vaddblueborder}{RGB}{88,124,166}
\usepackage{graphicx}
\graphicspath{{figures/}}
\usepackage{enumitem}
\usepackage{tikz}
\usetikzlibrary{arrows.meta, positioning, fit, backgrounds, calc}
\usepackage{algorithm}
\usepackage{algpseudocode}
\usepackage{wrapfig}

\newcommand{\tyes}{\textcolor{gooddark}{\checkmark}}
\newcommand{\tno}{\textcolor{baddark}{\ensuremath{\times}}}
\usepackage{tabularx}
\usepackage{booktabs}
\usepackage[table]{xcolor}
\usepackage{array}

\newcolumntype{Y}{>{\centering\arraybackslash}X}
\newcolumntype{B}{>{\columncolor{emoeblue!8}\centering\arraybackslash}X}

\usepackage[most]{tcolorbox}

\newsavebox{\emoetabbox}
\newtcolorbox{resultbox}{colback=black!3, colframe=black!55, boxrule=0.6pt, arc=2pt, left=6pt, right=6pt, top=6pt, bottom=6pt, fontupper=\small}

\usepackage[most]{tcolorbox}

\newtcolorbox{findingbox}[1]{%
    enhanced,
    breakable,
    colback=emoeblue!8,
    colframe=emoeblue!8,
    boxrule=0pt,
    arc=3mm,
    outer arc=3mm,
    boxsep=0pt,
    left=12pt,
    right=12pt,
    top=10pt,
    bottom=10pt,
    before skip=8pt,
    after skip=8pt,
    fontupper=\itshape,
    before upper={%
        \noindent{\normalfont\bfseries #1}\quad
    },
}

\title{E-MoE: Enhanced Mixture-of-Experts for \\ Non-Factorized Diffusion Language Models}

\author{
{\normalsize\bfseries
Arseny Ivanov$^{1,2}$
\qquad
Alexander Kolesov$^{2,1}$
\qquad
Alexander Korotin$^{2,1}$
}
\\[0.45em]
\hspace{0.35em}%
{\normalsize\bfseries
 Ivan Oseledets$^{1,2}$
\hspace{4em}
Mikhail Goncharov$^{1}$
}
\\[0.65em]
\hspace{0.35em}%
{\small
$^{1}$AXXX, Moscow, Russia
\hspace{3em}
$^{2}$Applied AI Institute, Moscow, Russia
}
}

\iclrfinalcopy 

\begin{document}

\vspace*{-0.2in}
\maketitle

\vspace{-4mm}
\begin{abstract}
Masked diffusion models (MDMs) generate sequences by progressively unmasking several tokens per denoising step, but their reverse process is typically factorized over positions, limiting sample quality in the few-step regime where diffusion's speed advantage over autoregressive decoding matters most. A recent line of work introduces a continuous Gaussian latent, trained as a variational autoencoder, to capture correlations across positions, but such approaches are prone to posterior collapse, where the latent is silently ignored. We propose \textbf{Enhanced Mixture-of-Experts (E-MoE)}, which builds the reverse process as a mixture of factorized distributions over a \emph{discrete shared latent} given by the expert-routing decisions of a Mixture-of-Experts (MoE) backbone, without increasing active parameters over the factorized baseline. Across synthetic multi-modal benchmarks, binarized MNIST, and LM1B, E-MoE improves few-step generation over factorized baselines.
\end{abstract}

\vspace{-0.25cm}
\begin{figure}[h]
  \centering
  \includegraphics[width=0.95\linewidth]{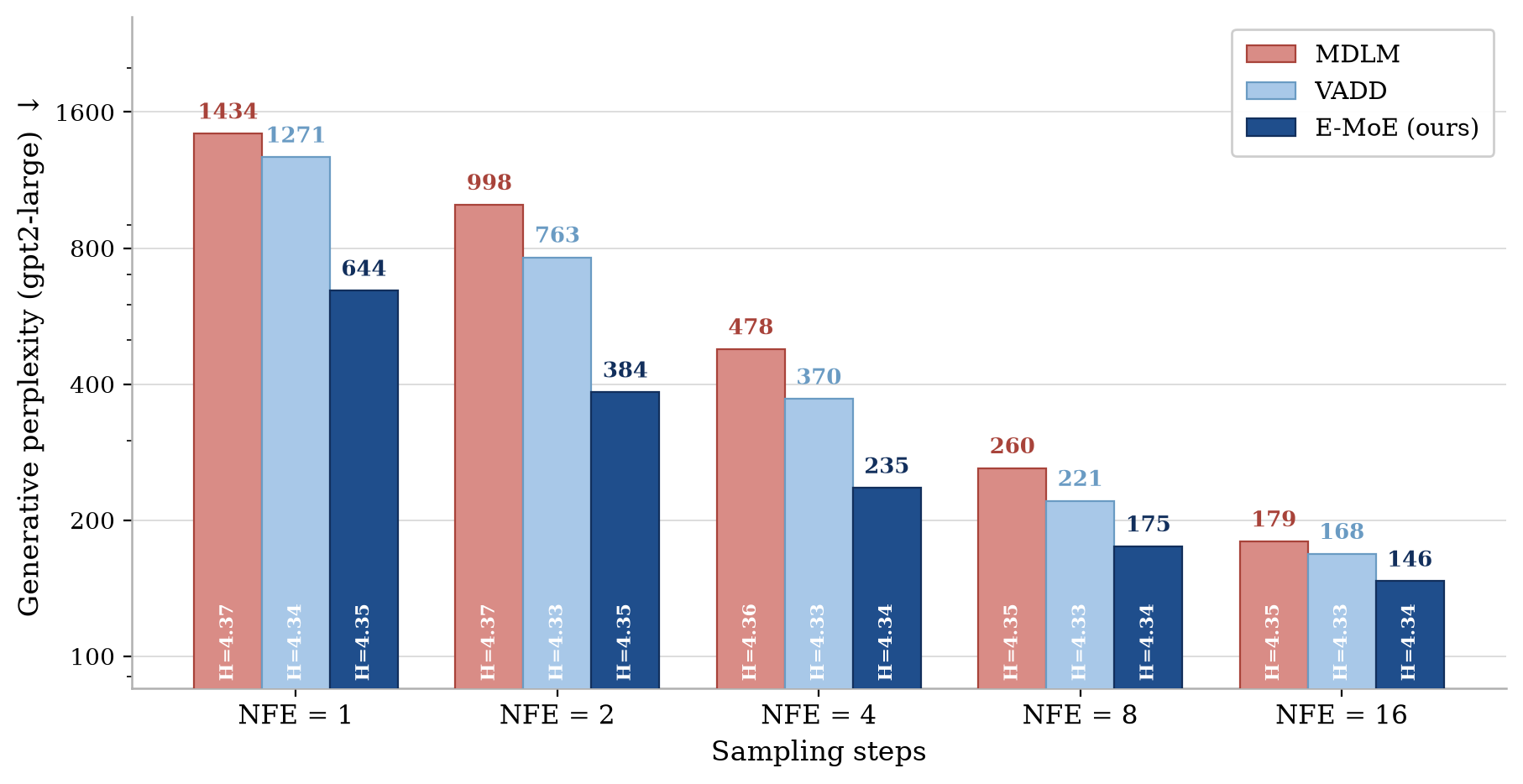}
  \vspace{-3mm}
  \caption{\textbf{Few-step generative perplexity on LM1B} (lower is better, $H$ = sample entropy).
  E-MoE gives markedly lower generative perplexity than both factorized MDLM~\citep{sahoo2024simple} and continuous-latent VADD~\citep{xie2026vadd} for all NFEs 1–16, at matched sample entropy.}
  \label{fig:teaser}
\end{figure}

\vspace{-1mm}
\section{Introduction}
\label{sec:intro}

\textbf{Autoregressive} (AR) language models~\citep{vaswani2017attention,brown2020language,yoo2026selfconditionedflowmaplanguage} are common and widespread approach for language modeling. They
factorize text along a fixed left-to-right order and generate one token per forward pass conditioned on previous tokens with causal attention. That ordering makes the likelihood exactly tractable, but it also makes decoding inherently sequential. Each token conditions on all previous ones, so a sequence of length $\text{L}$ costs $\text{L}$ forward passes that cannot be parallelized.

Recently, \textbf{Masked Diffusion Models} (MDM)~\citep{austin2021structured, lou2024sedd,sahoo2024simple,shi2024simplified} have suggested a different approach based on diffusion forward and backward  processes with $\text{T}$ steps. While the forward process gradually masks each token independently, the factorized model conditions on the whole sequence, with likelihood evaluated via bounds or estimators~\citep{ivanov2026tubetangentupperbound}, and predicts multiple tokens per forward pass with bidirectional attention at the each of $\text{T}$ steps of backward process. Thus, MDM recover a sequence of length $\text{L}$ costs $\text{T} < \text{L}$ forward passes, allowing for using of parallelization.

In practice, MDM still need many unmasking steps to produce coherent text. One of the main reason is the factorization of  a backward process. Although, model conditions on a whole sequence, it predicts an independent marginal distributions for every masked positions. Thereby, \textbf{the factorization causes the absence of correlations between  unmasked tokens obtained through a forward pass.}  Sampling several positions from these marginals treats them as conditionally independent and discards the dependencies between them~\citep{liu2025discretecopuladiffusion}, an error that grows with the number of tokens unmasked at once, precisely where parallel unmasking would pay off. 

One way to overcome the factorization error is to consideration of shared latent variable \citep{xie2026vadd}, that models hidden joint representation of current context and predicted sequence. Existing instantiations~\citep{xie2026vadd,shariatian2026ladd,zhou2026coevolutionary} take this latent to be \emph{continuous} and train it with an \emph{auxiliary} recognition network alongside MDM. The continuous structure of latent requires parametric prior, in practice a Gaussian, thereby not allowing for constructing more informative prior from data. The joint training of MDM with an additional recognition network implies the increasing of learnable parameters in the training stage. We ask whether such a latent can be obtained without restricting to fixed prior and joint training with an additional network. We positively answer to the question and model the reverse process over a \textbf{discrete} shared latent given by the expert-routing decisions of a mixture-of-experts (MoE) backbone~\citep{shazeer2017moe,fedus2022switch} - an architecture already used to scale MDM~\citep{llada2025, zhu2025lladamoesparsemoediffusion}. The same router, evaluated on the noisy and on the clean sequence, serves as both the generative prior and the training-time variational posterior, so no recognition model is added and no prior over the latent has to be designed. Our \textbf{contributions} are as follows:
 
\begin{itemize}[leftmargin=1.4em, itemsep=1pt, topsep=2pt]
  \item \textbf{Method.} We propose \textbf{Enhanced Mixture-of-Experts} (E-MoE) for overcoming the factorization error, reusing a mixture of \emph{experts} as a mixture of \emph{factorized
  distributions}: the per-token and per-layer routing decisions an MoE backbone already computes serve as the discrete shared latent, at no extra parameters and no extra inference cost
  (Section~\ref{sec:method}).
  \item \textbf{Training objective.} We derive the evidence lower bound (ELBO) for this mixture-based reverse process, together with a tractable bound on its latent KL that reduces to a per-layer and per-token quantity computed from the router itself (Section~\ref{sec:method}).
  \item \textbf{Empirical results.} On synthetic multi-modal data, binarized MNIST and LM1B, E-MoE improves few-step generation over both a factorized baseline and a continuous-latent one at matched sample entropy, and its latent remains in use throughout training without warmup schedules or auxiliary losses (Section~\ref{sec:experiments}).
\end{itemize}

\vspace{-2mm}
\section{Background}
\label{sec:background}

\vspace{-1mm}
\textbf{Notation.} We denote a $L$-token sequence $\textbf{x}:=(\textbf{x}^1,\dots,\textbf{x}^L)$, where $\textbf{x}^{\ell}:= \textbf{1}_{k}$ is a 'one-hot' column vector with $\text{K}$ positions and non-zero entry at $k$-th position. In terms of language modeling, $\text{K}$ is a vocabulary size and the vocabulary is the set $\mathcal{V}:= \{   \textbf{1}_{k}\}_{k=1}^{\text{K}}$, where 'one-hot' vector with non-zero entry at $\text{K}$-th position is referred to as special mask token $\textbf{m}$. Also, we introduce a masked subset of sequence $\textbf{x}$ as $\mathcal{M}(\textbf{x}): = \{ \textbf{x}^{\ell} : \textbf{x}^{\ell}= \textbf{1}_{\text{K}}\}_{\ell=1}^{L}$. Assuming the sequences $\textbf{x}_{0}$  are independent and sampled from a distribution $p_{data}(\textbf{x}_{0})$, we train $p_{\theta}(\textbf{x}_{0})$ to approximates it. We define $\text{Cat}(\cdot;\pi)$ as a categorical distribution over $\text{K}$ positions with corresponding probabilities given $\pi \in \triangle^{\text{K}}$, where $\triangle^{\text{K}}$ is the $\text{K}$-simplex. In particular, we introduce singular distribution $\pi_{\textbf{m}}:= \textbf{1}_{\text{K}}$.

\textbf{Autoregressive} (AR) language models use a sequential factorization $p_\theta(\textbf{x}_{0}) = \prod_{\ell=1}^{L} p_\theta(\textbf{x}_{0}^\ell \mid \textbf{x}_{0}^{<\ell})$ that lies at the heart of their model parameterized by a causal
attention \citep{vaswani2017attention}. However, sequential decoding is a bottleneck, so generation of $\text{L}$ tokens requires $\text{L}$ forward passes.

\textbf{Masked Diffusion Models} (MDM) ~\citep{sahoo2024simple, shi2024simplified, ou2025your} replace left-to-right decoding by a diffusion dynamics with forward and backward processes. The forward process gradually corrupts sequence $\textbf{x}_{s}$ to more masked  $\textbf{x}_{t}$, being factorized over all $\text{L}$ tokens independently:
\begin{equation}
\label{eq:mdlm_fwd}
    q(\textbf{x}_{t}|\textbf{x}_{s}) = \prod_{\ell=1}^{L} q(\textbf{x}_{t}^{\ell}|\textbf{x}_{s}^{\ell} ) = \prod_{\ell=1}^{L}\text{Cat}(\textbf{x}_{t}^{\ell}; \alpha_{t|s}\textbf{x}_{s}^{\ell} + (1 - \alpha_{t|s}) \textbf{m} ),
\end{equation}
where $0 \leq s \leq t \leq T$ and $\alpha_{t|s}:=\alpha_{t}/\alpha_{s}$ is  the ratio of decreasing schedules \citep[MDLM]{sahoo2024simple}. According to ~\citep{austin2021structured}, the forward process has the related reverse is given by:
\begin{equation}
\label{eq:mdlm_back_posterior}
 q(\textbf{x}_{s}|\textbf{x}_{t}, \textbf{x}_{0}) = \prod_{\ell=1}^{L} q(\textbf{x}_{s}^{\ell}|\textbf{x}_{t},  \textbf{x}_0^\ell) = \prod_{\ell=1}^{L} 
\begin{cases} 
\text{Cat}(\textbf{x}_{s}^\ell; \textbf{x}_t^\ell) & \textbf{x}_t^\ell \neq \mathbf{m}, \\[8pt]
\text{Cat}\left(\textbf{x}_{s}^\ell; \dfrac{(1-\alpha_s)\textbf{m} + (\alpha_s - \alpha_t)\textbf{x}_{0}^\ell}{1-\alpha_t}\right) & \textbf{x}_t^\ell = \mathbf{m}.
\end{cases}
\end{equation}

\begin{figure}[t]
\centering
\includegraphics[width=\linewidth]{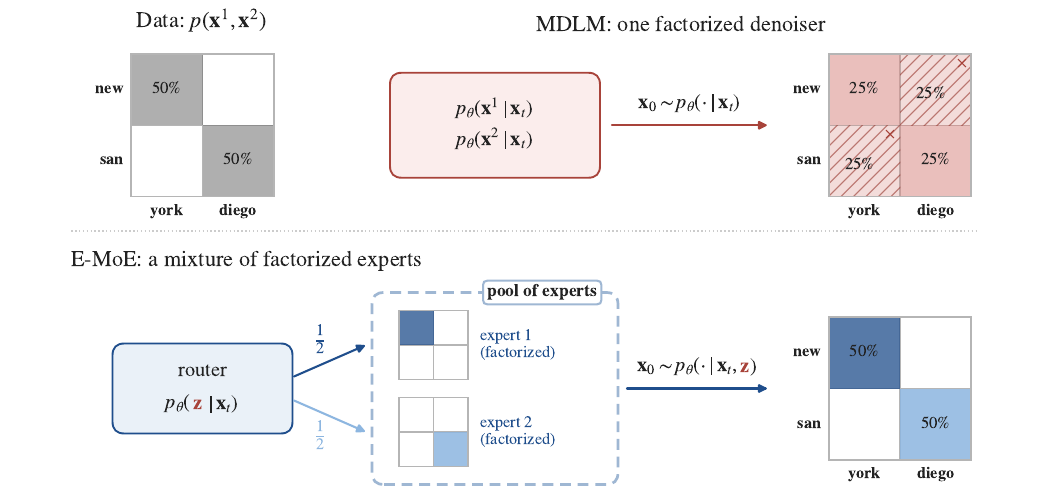}
\vspace{-4mm}
\caption{\textbf{Factorization error and mixture over expert routes.}
\textbf{Top:} a factorized reverse process can produce spurious pairings when multiple tokens are sampled at once.
\textbf{Bottom:} the routing code
$\mathbf{z}=\{\mathbf{z}_d\}_{d=1}^{D}$, with
$\mathbf{z}_d=(\mathbf{z}_d^\ell\}_{\ell=1}^{L}$,
specifies the expert assignment for every position across layers.
}
\label{fig:mixture-rank}
\end{figure}

The same construction (\ref{eq:mdlm_back_posterior}) is used for modeling a reverse process $p_{\theta}(\textbf{x}_{s}|\textbf{x}_{t})$, substituting not known a clean token   $\textbf{x}_{0}^\ell$  by a  model's estimation $\textbf{x}_\theta ^\ell\sim \mu_\theta^\ell(\textbf{x}_t,t) $  with bidirectional attention over the whole  $\textbf{x}_{t}$. The trained reverse process unmasks sequence from $\textbf{x}_{t}$ to less masked $\textbf{x}_{s}$, being parameterized by:
\begin{equation}
\label{eq:factor}
    p_{\theta}(\textbf{x}_{s}|\textbf{x}_{t}) = \prod_{\ell=1}^{L} p_{\theta}(\textbf{x}_{s}^{\ell}|\textbf{x}_{t}) = \prod_{\ell=1}^{L}  q(\textbf{x}_{s}^{\ell}|\textbf{x}_{t},  \textbf{x}_\theta^\ell).
\end{equation}
This factorization (\ref{eq:factor}) makes all positions easy to sample in parallel, allowing for sampling  $L$ tokens in $T < L$ forward passes. Nevertheless, it  creates the main \textbf{modeling weakness}: tokens sampled at the same denoising step $\textbf{x}_{s}^{\ell}$ and $\textbf{x}_{s}^{\ell'}$ are not explicitly correlated. This issue means that $p_\theta(\textbf{x}_s \mid \textbf{x}_t)$, being a product of token-wise marginals, is a rank-one approximation of the true joint $p_{data}(\textbf{x}_0)$, and therefore cannot represent any dependence between the revealed tokens, see Fig.~(\ref{fig:mixture-rank}).

\textbf{Mixture of marginals}. One of the possible ways to overcome the weakness is to consider a shared latent variable that models connection between $\textbf{x}_{s}$ and $\textbf{x}_{t}$, guiding the model to $p_{data}$ \citep[VADD]{xie2026vadd}. Then,  a \textbf{non-factorized} reverse process is defined by a mixture of factorized conditionals as:
\begin{equation}
\label{eq:cont_marginalization}
  p_\theta(\textbf{x}_s \mid \textbf{x}_t) = \int_{\textbf{z}} p_\theta(\textbf{x}_s \mid \textbf{x}_t, \textbf{z})\,p(\textbf{z})\,d\textbf{z},  
\end{equation}
where $p(\textbf{z})$ is a prior distribution with continuous  latent variable $\textbf{z} \in \mathbb{R}^{u}$. \textbf{The main idea of this approach} is to give an opportunity for modeling $p_{\theta}(\textbf{x}_s \mid \textbf{x}_t, \textbf{z})$ as multiplication of factorized distributions $ p_{\theta}(\textbf{x}_s^{\ell} \mid \textbf{x}_t, \textbf{z})$, while $p_\theta(\textbf{x}_s \mid \textbf{x}_t)$ remains \textbf{non-factorized} by integrating over $\textbf{z}$. VADD  uses the standard Gaussian distribution $\mathcal{N}(0,I_{u \times u})$ for prior   and parameterize reverse process as:
\begin{equation*}
\label{eq:vadd-forward}
\begin{aligned}
p_{\theta}(\textbf{x}_{s}|\textbf{x}_{t})
&= \mathbb{E}_{p(\textbf{z})}\prod_{\ell=1}^{L}
p_\theta(\textbf{x}_s^{\ell} \mid \textbf{x}_t , \textbf{z}) = \mathbb{E}_{p(\textbf{z})}\prod_{\ell=1}^{L}
\begin{cases}
\text{Cat}(\textbf{x}_s^\ell; \textbf{x}_t^\ell),
& \textbf{x}_t^\ell \neq \textbf{m}; \\[8pt]
\text{Cat}\!\left(\textbf{x}_s^\ell;
\dfrac{(1-\alpha_s)\textbf{m}+(\alpha_s-\alpha_t)\textbf{x}_{\theta}^{\ell}}{1-\alpha_t}\right),
& \textbf{x}_t^\ell = \textbf{m}.
\end{cases}
\end{aligned}
\end{equation*}
where $\textbf{x}_{\theta}^{\ell} \sim \mu_{\theta}^\ell(\textbf{x}_t, \textbf{z}, t)$ is the estimation of unknown clean token by the model. The training process of $\mu_{\theta}(\textbf{x}_t, \textbf{z}, t)$ follows to ELBO optimization as in MDLM. Besides, VADD requires training of autoencoder \citep[VAE]{kingma2013auto} for learning appropriate shared latent $z$ by   minimizing Kullback-Leibler (KL) divergence between fixed Gaussian prior $p(\textbf{z})$ and learnable posterior $p_{\phi}(\textbf{z}|\textbf{x}_{0},\textbf{x}_{t})$. Thus, the total loss function for VADD  $\mathcal{L}_{\text{vadd}}(\mathbf{x}_{0}, \theta, \phi, \lambda):= \mathcal{L}_{\text{vadd}}$  is defined as:
\begin{equation*}
\label{eq:elbo_vadd}
\begin{aligned}
\mathcal{L}_{\text{vadd}}
&:= \mathbb{E}_{q(\cdot|\mathbf{x}_{0})}
\mathbb{E}_{p_{\phi}(\cdot|\mathbf{x}_{0},\mathbf{x}_{t}),\, t \sim [0,1]}
\Bigg[
\underbrace{\sum_{\ell \in \mathcal{M}(\mathbf{x}_t)}
\frac{-\alpha_t'}{1 - \alpha_t}
\log p_{\theta}(\mathbf{x}^\ell_{0}| \mathbf{x}_{t},\mathbf{z})}_{\text{ELBO}}
+ \lambda \underbrace{\log\!\left(
\frac{p_{\phi}(\mathbf{z}|\mathbf{x}_{0},\mathbf{x}_{t})}{p(\mathbf{z})}
\right)}_{\text{KL}(p(\textbf{z})||p_{\phi}(\textbf{z}|\textbf{x}_{0},\textbf{x}_{t}))}
\Bigg].
\end{aligned}
\end{equation*}
However, it is difficult to find appropriate prior for data in continuous case, without restricting to a Gaussian distribution and additional VAE model on training. To overcome this issue, we consider \textbf{discrete} latent space and special MoE architecture in the next section that allow to sort it out.

\vspace{-2mm}
\section{Method}
\label{sec:method}

We build the reverse process over a discrete shared latent. We first state the upper bound on the negative ELBO for such a process (\S\ref{sec:discrete-latent}). Then we realize the latent as the routing decisions an MoE backbone already computes (\S\ref{sec:moe_param}), and turn the bound into a training objective and a sampler (\S\ref{sec:train}).

\subsection{Upper Bound with discrete shared latent}
\label{sec:discrete-latent}

Although, MDLM and VADD follow to the consideration of continuous time $t \in [0,1]$, we observe discrete time grid with time step $t_i = i/\text{T}$ in our approach, where $\text{T}$ is number of diffusion steps. Moving to our method for overcoming of the factorization issue, we not only consider a \textbf{discrete} shared latent $\textbf{z}$ instead of a continuous, but we also condition the prior $p_{\theta}(\textbf{z}|\textbf{x}_{t_i})$ on the data $\textbf{x}_{t_i}$ and  learn it to be more \textbf{appropriate} for the data, rather than restricting it to a known distribution. Regardless of the changes, $p_{\theta}(\textbf{x}_{t_s}|\textbf{x}_{t_i})$ is still  the mixture of factorized marginals, but with a sum as:
\begin{equation}
  p_\theta(\textbf{x}_{t_s} \mid \textbf{x}_{t_i})
  =
  \sum_{\textbf{z}}
  p_\theta(\textbf{z} \mid \textbf{x}_{t_i})\,
  p_\theta(\textbf{x}_{t_s} \mid \textbf{x}_{t_i},\textbf{z}),
  \qquad
  p_\theta(\textbf{x}_{t_s} \mid \textbf{x}_{t_i},\textbf{z})
  =
  \prod_{\ell=1}^{L}
  q\!\left(\textbf{x}_{t_s}^\ell \mid \textbf{x}_{t_i},
  \textbf{x}_{\theta}^{\ell}\right),
  \label{eq:emoe-mixture}
\end{equation}
where $ \textbf{x}_{\theta}^{\ell}$ is parametric approximation of clean token $\textbf{x}_{0}^{\ell}$. Furthermore, using this form of the prior, we derive an upper bound for negative log-likelihood  $\mathbb{E}_{p_{\text{data}}(\textbf{x}_{0})}[ - \log p_{\theta}(\textbf{x}_{0})]$, that is also used as  loss function for our method and postulated  below with the corresponding proof in Appendix \ref{app:prop2}:
 
\begin{findingbox}{Proposition 3.1}
\label{text_prop} Let $i \in \overline{1,T}$ is a number of diffusion step and $t_i:= \frac{i}{T}$ is a  $i$-th time step. Let $p_{\text{data}}(\textbf{x}_{0})$ is a data distribution of $L$-length sequences and $p_{\theta}(\textbf{x}_{0})$ is its parametric estimation. Let $\textbf{x}_{\theta} \sim \mu_{\theta}(\textbf{x}_{t_i}, t_{i})$ is a estimation of clean $\textbf{x}_{0}$ and $\mathcal{M}(\textbf{x}_t)$ is a masked subset of $\textbf{x}_t$. Let $q_{\theta}(\textbf{z}|\textbf{x}_{t_i}, \textbf{x}_{0})$ and $p_{\theta} (\textbf{z}|\textbf{x}_{t_i})$ are  posterior and prior distributions at $i$-th step. \\

Then, the upper bound of negative log-likelihood $\mathbb{E}_{p_{\text{data}(\textbf{x}_{0})}}[ - \log p_{\theta}(\textbf{x}_{0})]$ is given by:

\begin{equation}
\label{eq:our_loss}
\mathbb{E}_{p_{data}(\textbf{x}_{0})} \frac{1}{\text{T}} \sum_{i=1}^{\text{T}} \mathbb{E}_{q(\textbf{x}_{t_i}|\textbf{x}_{0})}\mathbb{E}_{q_{\theta}(\textbf{z}|\textbf{x}_{t_i},\textbf{x}_{0})} [\frac{-1}{t_i}\sum_{\ell \in \mathcal{M}(\textbf{x}_t)} \log p_{\theta}(\textbf{x}_{0}^{\ell}|\textbf{x}^{\ell}_{t_i},\textbf{z}) + \text{T}\log\frac{q_{\theta}(\textbf{z}|\textbf{x}_{t_i},\textbf{x}_{0})}{p_{\theta}(\textbf{z}|\textbf{x}_{t_i})}]
 \end{equation}
\end{findingbox}

\subsection{Mixture-of-Experts parameterization}
\label{sec:moe_param}

\begin{figure}[t]
\centering
\resizebox{0.98\linewidth}{!}{%
\begin{tikzpicture}[
  >=Latex,
  banner/.style={rounded corners=3pt, minimum height=8mm, minimum width=62mm, align=center, font=\bfseries\normalsize, text=white},
  plainbox/.style={draw=black!70, fill=white, rounded corners=2pt, minimum height=7mm, align=center, font=\small},
  tintbox/.style={draw=black!70, rounded corners=2pt, minimum height=7mm, align=center, font=\bfseries\small},
  cell/.style={draw=black!70, minimum width=6mm, minimum height=6mm, font=\small},
  tok/.style={rounded corners=2.5pt, line width=0.8pt, minimum width=9mm,
              minimum height=5.6mm, inner xsep=2pt, font=\bfseries\scriptsize},
  wtokL/.style={tok, draw=emoeblueborder, fill=emoeblue!10},
  wtokR/.style={tok, draw=mdlmredborder, fill=mdlmred!12},
  wtokO/.style={tok, draw=black!70, fill=white},
  mtok/.style={tok, draw=black!55, fill=black!10, text=black!65},
  plusnode/.style={draw=black!70, fill=white, circle, minimum size=5mm, font=\bfseries\small, inner sep=0pt},
  lbl/.style={font=\small, align=center, fill=white, inner sep=1pt},
  small lbl/.style={font=\footnotesize, align=center},
  conn lbl/.style={font=\small, align=center, inner sep=1.5pt}
]

\draw[emoeblueborder, dash pattern=on 3pt off 2pt, line width=0.9pt, rounded corners=8pt, fill=emoeblue!4]
  (0.00,3.3) rectangle (7.10,10.9);
\node[banner, fill=emoeblueborder] at (3.55,10.25) {NOISY PASS $\rightarrow$ PRIOR};

\node[small lbl] at (0.39,9.15) {$\textbf{x}_{_i}$};
\node[wtokL] (xt1) at (1.124,9.15) {THE};
\node[mtok,  right=2pt of xt1] (xt2) {[M]};
\node[wtokL, right=2pt of xt2] (xt3) {WILL};
\node[mtok,  right=2pt of xt3] (xt4) {[M]};
\node[wtokL, right=2pt of xt4] (xt5) {AT};
\node[mtok,  right=2pt of xt5] (xt6) {[M]};

\node[plainbox, minimum width=58mm] (blk1) at (3.55,8.1) {Norm $\to$ Self-Attention $\to$ Norm};
\draw[->] (3.55,8.70) -- (blk1);

\node[tintbox, fill=emoeblue!20, minimum width=30mm] (gateL) at (3.55,6.55) {Router};
\draw[->] (blk1) -- (gateL);

\node[small lbl] at (3.55,5.72) {prior $p_\theta(\textbf{z}_{t_i}\mid\textbf{x}_{t_i})$};
\draw[black!25, line width=0.6pt] (1.97,3.85) -- (5.13,3.85);
\foreach \i/\h/\hi in {0/0.59/0,1/1.28/1,2/0.74/0,3/1.47/1,4/0.41/0}{
  \pgfmathsetmacro{\x}{2.07+\i*0.62}
  \ifnum\hi=1
    \draw[draw=emoeblueborder, line width=1.1pt, fill=emoeblue!55] (\x,3.85) rectangle ++(0.48,\h);
  \else
    \draw[draw=black!60, fill=black!8] (\x,3.85) rectangle ++(0.48,\h);
  \fi
}

\draw[mdlmredborder, dash pattern=on 3pt off 2pt, line width=0.9pt, rounded corners=8pt, fill=mdlmred!6]
  (9.55,3.3) rectangle (16.65,10.9);
\node[banner, fill=mdlmredborder] at (13.1,10.25) {CLEAN PASS $\rightarrow$ POSTERIOR};

\node[small lbl] at (9.94,9.15) {$\textbf{x}_0$};
\node[wtokR] (xc1) at (10.674,9.15) {THE};
\node[wtokR, right=2pt of xc1] (xc2) {SUN};
\node[wtokR, right=2pt of xc2] (xc3) {WILL};
\node[wtokR, right=2pt of xc3] (xc4) {RISE};
\node[wtokR, right=2pt of xc4] (xc5) {AT};
\node[wtokR, right=2pt of xc5] (xc6) {DAWN};

\node[plainbox, minimum width=62mm] (blk2) at (13.1,8.1) {Norm $\to$ Self-Attention $\to$ Norm \scriptsize(shared weights)};
\draw[->] (13.1,8.70) -- (blk2);

\node[tintbox, fill=mdlmred!25, minimum width=30mm] (gateR) at (13.1,6.55) {Router};
\draw[->] (blk2) -- (gateR);

\node[small lbl] at (13.1,5.72) {target $q_\theta(\textbf{z}_{t_i}\mid\textbf{x}_{t_i},\textbf{x}_0)$};
\draw[black!25, line width=0.6pt] (11.52,3.85) -- (14.68,3.85);
\foreach \i/\h/\hi in {0/1.47/1,1/0.54/0,2/0.68/0,3/0.38/0,4/0.86/0}{
  \pgfmathsetmacro{\x}{11.62+\i*0.62}
  \ifnum\hi=1
    \draw[draw=mdlmredborder, line width=1.1pt, fill=mdlmred!65] (\x,3.85) rectangle ++(0.48,\h);
  \else
    \draw[draw=black!60, fill=black!8] (\x,3.85) rectangle ++(0.48,\h);
  \fi
}

\draw[<->, dashed, black!55] (gateL.east) -- (gateR.west);
\node[conn lbl, text=black!55] at (8.325,6.80) {same weights};
\draw[<->, dashed, black!70] (5.4,4.60) -- (11.25,4.60);
\node[conn lbl, text=black!70] at (8.325,4.85) {$\mathrm{KL}(q_\theta\Vert p_\theta)$};

\coordinate (cx) at (8.325,2.2);
\node[draw=black!70, fill=white, circle, minimum size=6mm, inner sep=0pt] (junc) at (cx) {};
\draw[->, dotted, thick, emoeblueborder] (3.55,3.75) -- (3.55,2.2) -- (junc);
\draw[->, thick, mdlmredborder] (13.1,3.75) -- (13.1,2.2) -- (junc);

\node[small lbl, text=emoeblueborder, align=center] at (5.6,2.65) {inference:\\$\textbf{z}_{t_i} \sim p_\theta$};
\node[small lbl, text=mdlmredborder, align=center] at (11.05,2.65) {training:\\$\textbf{z}_{t_i} \sim q_\theta$};

\node[plainbox, minimum width=19mm] (e1) at (5.10,1.3) {Expert 1};
\node[plainbox, minimum width=19mm] (e2) at (7.70,1.3) {Expert 2};
\node[small lbl, minimum height=7mm] (edots) at (9.55,1.3) {$\cdots$};
\node[plainbox, minimum width=19mm] (en) at (11.55,1.3) {Expert $E$};
\foreach \e in {e1,e2,en}{ \draw[->] (junc) -- (\e); }

\node[plusnode] (p2) at (8.325,0.15) {+};
\draw[->] (e1) |- (p2);
\draw[->] (e2) -- (p2);
\draw[->] (en) |- (p2);

\node[small lbl] at (5.17,-0.55) {$\hat{\textbf{x}}_{\theta,\textbf{z}_t}$};
\node[wtokO] (o1) at (5.899,-0.55) {THE};
\node[wtokO, right=2pt of o1] (o2) {SUN};
\node[wtokO, right=2pt of o2] (o3) {WILL};
\node[wtokO, right=2pt of o3] (o4) {RISE};
\node[wtokO, right=2pt of o4] (o5) {AT};
\node[wtokO, right=2pt of o5] (o6) {DAWN};
\draw[->] (p2) -- (8.325,-0.25);

\end{tikzpicture}%
}
\caption{\textbf{Training stage:} we model prior $p_{\theta}(\textbf{z}_{t_i}|\textbf{x}_{t_i})$ and posterior $q_{\theta}(\textbf{z}_{t_i}|\textbf{x}_{t_i},\textbf{x}_{0})$ by the \textbf{same} router, unlike VADD. The router $\rho_{\theta}(\cdot,t_i)$ takes as input a noisy sample $\textbf{x}_{t_i}$ to approximate $p_{\theta}$  in the noisy pass, and  a concatenation of $\textbf{x}_{t_i}$ and $\textbf{x}_{0}$ for $q_{\theta}$ in the clean pass. Then, the experts return predictions for each masked  $\textbf{x}_{t_i}^{\ell}$ using appropriate $\textbf{z}_{t_i}$.  \textbf{Inference stage:} we use only the  noisy pass (Section~\ref{sec:moe_param}).}
\label{fig:method-diagram}
\end{figure}
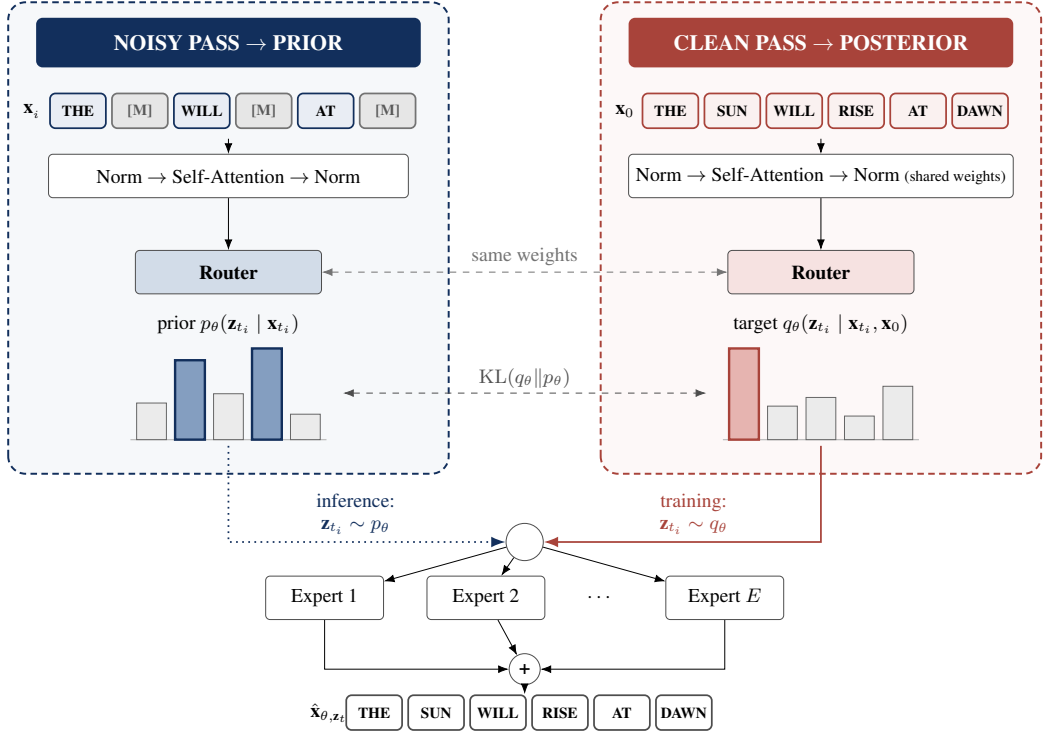

To avoid a learning of additional models as VAE in VADD during the training stage, we instead use the routing decisions of a mixture-of-experts (MoE) model~\citep{shazeer2017moe,fedus2022switch}. The main advantage of MoE architecture that latent is already inserted inside of the model by routing mechanism and is not an extra continuous vector required an additional training. This latent is the discrete choice of which expert, or sparse mixture
of experts, should explain the  unmasking step.

MoE model is composed of $\text{E}$ experts with $\text{D}$ layers, where each  $e$-th expert $\mu_{\theta, e}^{\ell}(\textbf{x}_{t_i},t_i) \in \triangle^{\text{K}}$ estimates $\textbf{x}_{0}^{\ell}$. Besides, there is a prior routing mechanism $\rho_{\theta,d}^{\ell}(\textbf{x}_{t_i},t_i)\in\triangle^{\text{E}}$ that assigns which expert processes $l$-th token. Since each expert consists of $\text{D}$ layers, we model a latent code $\textbf{z}$ for  a sequence $\textbf{x}_{t_i}$ as $\textbf{z}:= \{ \textbf{z}_{d}\}_{d=1}^{D}$ with $\textbf{z}_{d}:= (\textbf{z}_{d}^{1},...,\textbf{z}_{d}^{L}) $, where $\textbf{z}_{d}^{\ell}$ is a latent code at $d$-th layer, meaning which expert is assigned by a routing mechanism for $\ell$ position in $\textbf{x}_{t_i}$. Since the prior distribution $q_{\theta}(\textbf{z}|\textbf{x}_{t_i} )$ might be rewritten as $q_{\theta}( \{\textbf{z}_{1}^{1},...,\textbf{z}_{1}^{L}, ..., \textbf{z}_{D}^{L}\}|\textbf{x}_{t_i})$, we   assume dependence between current and previous latent codes at the same positions  and model $q_{\theta}(\textbf{z}^{\ell}|\textbf{x}_{t_i} )$ as  $q_{\theta}(\textbf{z}_{D}^{\ell}|\textbf{z}_{<D}^{\ell}, \textbf{x}_{t_i} ) q_{\theta}(\textbf{z}_{D-1}^{\ell}|\textbf{z}_{<D-1}^{\ell}, \textbf{x}_{t_i} )\cdot...\cdot
q_{\theta}(\textbf{z}_{1}^{L}|\textbf{x}_{t_i} )$, where $\textbf{z}_{<d}^{\ell}:= \{ \textbf{z}_{d-1}^{\ell},...,\textbf{z}_{1}^{\ell} \}$ are latent codes for all previous layers of the same positions. To provide related target to this prior, we introduce the posterior routing $\rho_{\theta,d}^{\ell}(\textbf{x}_{t_i},\textbf{x}_{0},t_i)\in\triangle^{\text{E}}$ that assigns experts for each $\textbf{x}_{0}^\ell$. Thus, using these routing decisions, we derive posterior and prior distributions via per-layer and per-token factorization as:

\vspace{-6mm}
\begin{equation}
\label{eq:prior_post}
   \underbrace{p_{\theta}(\mathbf{z}|\mathbf{x}_{t_i},\mathbf{x}_{0}) = \prod_{d=1}^{D}\prod_{\ell=1}^{L} p_{\theta}(\mathbf{z}^{\ell}_{d}|\mathbf{z}^{\ell}_{<d},\mathbf{x}_{t_i},\mathbf{x}_{0})}_{\text{posterior}}, \quad
   \underbrace{q_{\theta}(\mathbf{z}|\mathbf{x}_{t_i} ) = \prod_{d=1}^{D}\prod_{\ell=1}^{L} q_{\theta}(\mathbf{z}^{\ell}_{d}|\mathbf{z}^{\ell}_{<d},\mathbf{x}_{t_i})}_{\text{prior}}.
\end{equation}
\vspace{-4mm}

Since prior and posterior distributions has the equal factorization (\ref{eq:prior_post}), KL divergence between them in (\ref{eq:our_loss}) leads to a sum of KL per-layer and per-token. While prior is a categorical distribution that router outputs conditioned on $\textbf{x}_{\theta}$ at $d$-th layer, posterior is another categorical  obtained by router with $\textbf{x}_{0}$. Thus, the KL  aligns experts assigned by router $\rho_{\theta,d}^{\ell}(\textbf{x}_{t_i},\textbf{x}_{0},t_i)$ for a clean sequence $\textbf{x}_{0}$ and   $\rho_{\theta,d}^{\ell}(\textbf{x}_{t_i},t_i)$ for its estimation at $l$-th position and $d$-th layer.

Figure~\ref{fig:method-diagram} sums up our method. During the \textbf{training} stage, our model requires two forward passes. The first pass computes posterior distribution $p_{\theta}(\mathbf{z}|\mathbf{x}_{t_i},\mathbf{x}_{0})$ via posterior routing-decision mechanism and provides an estimation  $\mathbf{x}_{\theta}$ by experts. The second pass computes related prior $q_{\theta}(\mathbf{z}|\mathbf{x}_{t_i})$. Importantly, prior and  posterior in (\ref{eq:prior_post}) use the \textbf{same} router parameters.  The main
training problem is therefore to align the route selected when the clean sample
is available with the route selected from the corrupted sequence alone. During the \textbf{sampling} stage, our model use only prior $p_{\theta}(\mathbf{z}|\mathbf{x}_{t_i})$.

\subsection{Training and sampling}
\label{sec:train}
 
\textbf{Loss function.} As it mentioned before, we use (\ref{eq:our_loss}) as the loss function for our method's training. Here we only write
this objective in the MoE parameterization  of Section~\ref{sec:moe_param}. For a fixed diffusion step $i$, the left term in (\ref{eq:our_loss})
uses the experts through the routed mechanism as:
\[
  \textbf{x}_{\theta}^{\ell}\sim
  \mu_{\theta, e}^{\ell}(\textbf{x}_{t_i},\textbf{z},t_i),
  \qquad
  p_{\theta}(\textbf{x}^{\ell}\mid\textbf{x}_{t_i}^{\ell},\textbf{z})
  :=
  \mathrm{Cat}\!\left(
  \textbf{x}^{\ell};
  \mu_{\theta, e}^{\ell}(\textbf{x}_{t_i},\textbf{z},t_i)
  \right).
\]
Thus, conditioned on a sampled route $\textbf{z}$, the selected experts provide
the token distributions that enter the left term in  (\ref{eq:our_loss}). The right term is the router-matching term. Using
the factorization from (\ref{eq:prior_post}), it becomes a sum of categorical KL
divergences over all layers and  positions, denoted as $\mathbf{KL}_{\text{route}}$:
\begin{equation}
\resizebox{0.97\linewidth}{!}{$\displaystyle
\mathbb{E}_{q_{\theta}}
\left[
\log
\frac{
q_{\theta}(\cdot\mid\textbf{x}_{t_i},\textbf{x}_{0})
}{
p_{\theta}(\cdot\mid\textbf{x}_{t_i})
}
\right] = 
\sum_{d, \ell}
\mathrm{KL}\!\left(
\mathrm{Cat}\!\left(\cdot;
\rho_{\theta,d}^{\ell}(\textbf{x}_{t_i},\textbf{x}_{0},t_i)\right)
\,\middle\|\,
\mathrm{Cat}\!\left(\cdot;
\rho_{\theta,d}^{\ell}(\textbf{x}_{t_i},t_i)\right)
\right):=\mathbf{KL}_{\text{route}}
$}
\label{eq:emoe-kl}
\end{equation}
Substituting  
the right  term in (\ref{eq:our_loss}) by  (\ref{eq:emoe-kl}), we give our training  objective  $\mathcal{L}_{\text{E-MoE}}:= \mathcal{L}_{\text{E-MoE}}(\theta, \textbf{x}_{0})$ as:
\begin{equation}
\begin{aligned}
\mathcal{L}_{\text{E-MoE}}
&:=
\mathbb{E}_{p_{\mathrm{data}}(\textbf{x}_{0}), t_i}
\mathbb{E}_{q(\textbf{x}_{t_i}\mid\textbf{x}_{0})}
\Bigg[\mathbb{E}_{q_{\theta}(\textbf{z}\mid\textbf{x}_{t_i},\textbf{x}_{0})}
\frac{-1}{t_i}
\sum_{\ell\in\mathcal{M}(\textbf{x}_{t_i})}
\log p_{\theta}(\textbf{x}_{0}^{\ell}\mid\textbf{x}_{t_i}^{\ell},\textbf{z})  
+ \text{T}\cdot\mathbf{KL}_{\text{route}}
\Bigg],
\end{aligned}
  \label{eq:emoe-objective}
\end{equation}
where $t_i$ is sampled from uniform  over integers $\overline{1,T}$. The training algorithm is described in Algo.~\ref{alg:emoe-training}.

\textbf{Gumbel-Softmax.} Since our shared latent $\textbf{z}$ is discrete, to provide learning and  sampling from  prior and posterior routings, we use Gumbel-Softmax trick  with straight-through estimator \citep{jang2016categorical} instead of non-differentiable argmax. In accordance with the trick, we draw i.i.d. Gumbel noise  $g^\ell_{d,e}$, compute logits of posterior $\rho^\ell_{\theta,d}(\mathbf{x}_{t_i},\mathbf{x}_0,t_i)$ and define relaxed routing  $\widetilde{\mathbf z}_d^\ell\in\Delta^E$ as:
\begin{equation}
  \widetilde{z}_{d}^{\ell}
  =
  \frac{
  \exp\!\left(
  \left(\log \rho_{\theta,d}^{\ell}(\textbf{x}_{t_i},\textbf{x}_{0},t_i)_e
  + g_{d,e}^{\ell}\right)/\tau
  \right)
  }{
  \sum_{j=1}^{\text{E}}
  \exp\!\left(
  \left(\log \rho_{\theta,d,j}^{\ell}(\textbf{x}_{t_i},\textbf{x}_{0},t_i)
  + g_{d,j}^{\ell}\right)/\tau
  \right)
  },
  \label{eq:gumbel-softmax}
\end{equation}

where $\tau>0$ is a temperature parameter controlling how close the relaxed
route is to a one-hot expert assignment. Having denoted selected top-k experts as 
$\mathcal{S}_d^\ell := \operatorname{TopK}(\widetilde{\mathbf z}_d^\ell,k)$, we renormalize them and sample a number of expert at $d$-th layer for $\textbf{x}_{t_i}^{\ell}$ as:

\begin{equation}
\textbf{z}^\ell_{d}
\sim
\frac{
    \widetilde z_{d,e}^\ell
    \mathbbm{1}[e\in\mathcal{S}_d^\ell]
}{
    \sum_{j\in\mathcal{S}_d^\ell}\widetilde z_{d,j}^\ell
}
\end{equation}

\begin{figure}[t]
\centering
\begin{minipage}[t]{0.485\textwidth}
\begin{algorithm}[H]
\footnotesize
\caption{Training}
\label{alg:emoe-training}
\begin{algorithmic}[1]
\Require Diffusion grid $t_i=i/T$, temperature $\tau$.
\State Initialize the experts and routers;
\Repeat
  \State Sample $\textbf{x}_{0}\sim p_{\mathrm{data}}$, $i\sim\mathrm{Unif}\{1,\dots,T\}$;
  \State Sample $\textbf{x}_{t_i}\sim q(\textbf{x}_{t_i}\mid\textbf{x}_{0})$;
  \Statex \quad\textbf{First pass} (posterior):
  \State Compute $q_{\theta}(\textbf{z}\mid\textbf{x}_{t_i},\textbf{x}_{0})$ in \eqref{eq:prior_post};
  \State Sample $\textbf{z}\sim q_{\theta}(\textbf{z}\mid\textbf{x}_{t_i},\textbf{x}_{0})$ by \eqref{eq:gumbel-softmax};
  \State Sample $\textbf{x}_{\theta} \sim \mu_{\theta(e)}(\textbf{x}_{t_i},\textbf{z},t_i)$ ;
  \Statex \quad\textbf{Second pass} (prior):
  \State Compute $p_{\theta}(\textbf{z}\mid\textbf{x}_{t_i})$ in \eqref{eq:prior_post};
  \State Update $\theta$ using \eqref{eq:emoe-objective};
\Until{convergence}
\end{algorithmic}
\end{algorithm}
\end{minipage}\hfill%
\begin{minipage}[t]{0.485\textwidth}
\begin{algorithm}[H]
\footnotesize
\caption{Sampling}
\label{alg:emoe-sampling}
\begin{algorithmic}[1]
\Require Diffusion grid $t_i=i/T$, sequence length $L$.
\State Initialize $\textbf{x}_{t_T}\gets(\textbf{m},\dots,\textbf{m})$;
\For{$i=T,\dots,1$}
  \Statex \quad\textit{Route selection (prior):}
  \State Sample $\textbf{z}\sim p_{\theta}(\textbf{z}\mid\textbf{x}_{t_i})$;
  \Statex \quad\textit{Reverse transition:}
  \State Obtain $p_{\theta}(\textbf{x}^{\ell}\mid\textbf{x}_{t_i}^{\ell},\textbf{z})$ from the experts;
  \State Sample $\textbf{x}_{t_{i-1}}\sim p_{\theta}(\cdot\mid\textbf{x}_{t_i},\textbf{z})$;
\EndFor
\State \Return $\textbf{x}_{t_0}$.
\Statex \vphantom{X}
\end{algorithmic}
\end{algorithm}
\end{minipage}
\end{figure}

\paragraph{\textbf{Sampling}.} During the inference stage, we sample latent from the prior $p_{\theta}(\textbf{z}\mid\textbf{x}_{t_i})$  and then compute  $\textbf{x}_{\theta}$  during the same forward pass, thereby one NFE requires one forward pass. See details in Algo. \ref{alg:emoe-sampling}.

\begin{table}[t]
\centering
\renewcommand{\arraystretch}{1.3}
\resizebox{\textwidth}{!}{%
\begin{tabular}{@{}l c c >{\columncolor{emoeblue!8}}c@{}}
\toprule
 & \textbf{Factorized} & \multicolumn{2}{c}{\textbf{Mixture of factorized distributions}} \\
\cmidrule(lr){2-2}\cmidrule(lr){3-4}
 & MDLM~\citep{sahoo2024simple} & VADD~\citep{xie2026vadd} & \textbf{E-MoE (ours)} (\S\ref{sec:method}) \\[2pt]
 & $\displaystyle\prod_{\ell}p_{\theta}(\textbf{x}^{\ell}\mid\textbf{x}_{t})$
 & $\displaystyle\int p(\textbf{z})\prod_{\ell}p_{\theta}(\textbf{x}^{\ell}\mid\textbf{x}_{t},\textbf{z})\,d\textbf{z}$
 & $\displaystyle\sum_{\textbf{z}}p_{\theta}(\textbf{z}\mid\textbf{x}_{t})\prod_{\ell}p_{\theta}(\textbf{x}^{\ell}\mid\textbf{x}_{t},\textbf{z})$ \\
\midrule
Latent space                       & ---   & $\mathbb{R}^{d}$                  & $\{1,\dots,\text{E}\}^{L\times D}$ \\
Latent is chosen                   & ---   & once per sequence                 & \textbf{per token and layer}       \\
Learned, data-dependent prior      & \tno  & \tno                              & \tyes                              \\
Posterior $q$ comes from              & ---   &  $q_{\phi}$ from VAE (additional net)               & $q_{\theta}$ from \textbf{the same net}          \\  
\midrule
No extra active parameters                & \tyes & \tno                              & \tyes                              \\
Forward-passes per training step           & $1$   & $2$                               & $2$                                \\
Forward-passes per sampling step           & $1$   & $1$                               & $1$                                \\
\bottomrule
\end{tabular}%
}
\vspace{-1mm}
\caption{A summary of the design choices behind E-MoE and its two baselines. VADD needs a continuous latent with a fixed Gaussian prior and a separate recognition network, while E-MoE reuses the routing decisions the backbone already makes, so it trains as simply as the factorized baseline.}
\label{tab:method-comparison}
\label{tab:method-comparison}
\end{table}

\section{Related Work}
\label{sec:related_work}

\textbf{Distillation of MDM.} One line of work keeps the factorized parameterization of MDLM~\citep{sahoo2024simple} and shortens the sampling trajectory instead. \textbf{SDTT}~\citep{deschenaux2025sdtt} distills MDMs into itself progressively, reducing the number of sampling steps by a factor of two at each stage. \textbf{DUO}~\citep{sahoo2025the, deschenaux2026the} derives a duality between uniform-state discrete and Gaussian diffusion and uses it to port consistency distillation to the discrete setting. \textbf{DiMO}~\citep{zhu2025di} matches the teacher in a single forward pass and \textbf{IDLM}~\citep{li2026idlm} trains the student adversarially in the inverse direction. However, these methods require a pre-trained model, whereas we train our model from scratch, therefore, these works are out of scope.


\textbf{Overcoming the factorization error.} We ask instead whether a model can express correlated steps natively. \textbf{DCD}~\citep{liu2025discretecopuladiffusion} augments the denoiser with a copula model that restores the joint structure among simultaneously denoised positions, and \textbf{CoDD}~\citep{li2026breaking} replaces the factorized output with a tractable correlated layer. Another group changes the process itself:
\textbf{ReDi}~\citep{yoo2026redi} re-couples the trajectories so as to reduce the conditional total correlation, and \textbf{FLDD}~\citep{bartosh2026forwardlearneddiscretediffusionlearning} learns the forward process with the same goal. However, these methods either incur significant computational overhead or rely on post-training fine-tuning. Closest to us is \textbf{VADD}~\citep{xie2026vadd}, which introduces a shared Gaussian latent. As Table~\ref{tab:method-comparison} makes precise, E-MoE keeps this mixture view, but its latent is the routing decision the backbone already computes: the prior is learned from data and no recognition network is needed.

\vspace{-2mm}
\section{Experiments}
\label{sec:experiments}


In this section, we evaluate E-MoE on three tasks: two-dimensional toy examples (Section~\ref{sec:toy}), pixel-level image generation (Section~\ref{sec:mnist}), and text generation (Section~\ref{sec:lm1b}). Throughout, we compare three models -- a factorized MDLM~\citep{sahoo2024simple} baseline, VADD~\citep{xie2026vadd}, and E-MoE -- \emph{matched in backbone size, optimizer and training budget, so any gain comes from the routing mechanism alone}. \underline{The experimental details} are in Appendix~\ref{app:exp-details}.

\subsection{Two-dimensional toy examples}
\label{sec:toy}

A factorized reverse process can only represent per-position marginals, so at low NFE it can combine independently-sampled, individually-plausible values into a joint sample that does not exist in the data. We test whether E-MoE's shared discrete latent fixes this on two synthetic 2-D densities where the failure is easy to see and measure.

\paragraph{Setup.} We generate $n=20{,}000$ points per density, discretized into $\mathcal{V}=50$ tokens per coordinate ($L=2$). \textbf{8-modes} places $M=8$ Gaussian clusters. A factorized sampler can combine two clusters' coordinates into a spurious mode that lies between them and matches neither. \textbf{Swiss-roll} instead places mass on a thin 1-D spiral. The analogous failure fills the area around the manifold. The experimental setup details are provided in Appendix~\ref{app:toy-details}.

\begin{figure}[h]
  \centering
  \includegraphics[width=\linewidth]{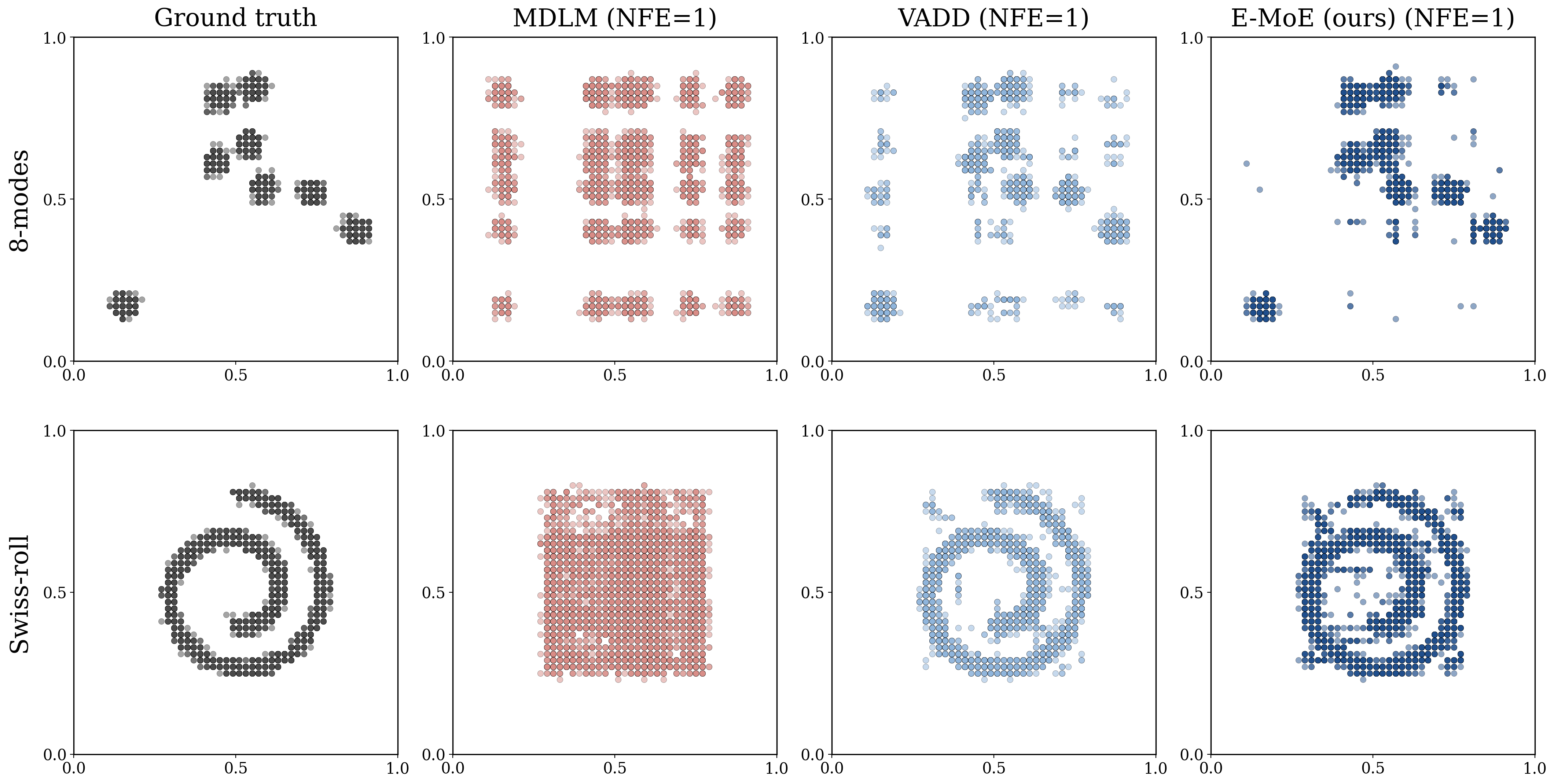}
  \vspace{-5mm}
  \caption{\textbf{Generation results on 2-D toy examples.} Ground truth and generations of MDLM~\citep{sahoo2024simple}, VADD~\citep{xie2026vadd}, and E-MoE at NFE~$=1$ on 8-modes and swiss-roll. Full sweeps over NFE~$\in\{1,2,8,32\}$ are in Appendix~\ref{app:toy-results}.}
  \label{fig:toy-summary}
\end{figure}

\paragraph{Results.}
Figure~\ref{fig:toy-summary} shows exactly the mode-averaging failure predicted above: at NFE~$=1$, MDLM's independently-sampled coordinates scatter across a blurred grid on 8-modes and fill the entire disk on Swiss-roll, while E-MoE stays concentrated on the eight true clusters and traces the spiral manifold.
Table~\ref{tab:toy} confirms this quantitatively: both VADD and E-MoE substantially improve validity over factorized MDLM in the few-step regime, with E-MoE consistently outperforming VADD on 8-modes and remaining competitive on Swiss-roll.
These toy experiments show that E-MoE's shared discrete latent captures cross-token correlation and \underline{breaks the factorization barrier}.

\begin{table}[h]
\caption{\textbf{Validity ($\uparrow$) of generated samples across sampling steps (NFE).}
Values are averaged over 3 seeds.
Best per column in \textbf{bold}.}
\vspace{3pt}
\label{tab:toy}
\centering
\small
\setlength{\tabcolsep}{4pt}
\resizebox{\textwidth}{!}{%
\begin{tabular}{lcccccccccccc}
\toprule
& \multicolumn{6}{c}{8-modes} & \multicolumn{6}{c}{Swiss-roll} \\
\cmidrule(lr){2-7}\cmidrule(lr){8-13}
Model
& NFE $=1$ & $2$ & $4$ & $8$ & $16$ & $32$
& NFE $=1$ & $2$ & $4$ & $8$ & $16$ & $32$ \\
\midrule

MDLM
& 37.7
& 68.1
& 84.2
& 90.9
& 94.9
& 97.0
& 49.3
& 73.1
& 86.5
& 91.9
& 95.2
& 96.9 \\

VADD
& 91.8
& 95.3
& 97.3
& 97.7
& 98.5
& 98.5
& \textbf{93.8}
& \textbf{95.9}
& \textbf{97.1}
& \textbf{97.6}
& 98.2
& \textbf{98.4} \\

E-MoE (ours)
& \textbf{94.4}
& \textbf{96.6}
& \textbf{97.7}
& \textbf{98.3}
& \textbf{98.6}
& \textbf{98.7}
& 90.3
& 92.6
& 95.7
& 96.4
& \textbf{98.6}
& 97.9 \\

\bottomrule
\end{tabular}%
}

\end{table}

\subsection{Pixel-level image generation}
\label{sec:mnist}

We next evaluate E-MoE on binarized MNIST, following the VADD evaluation setup~\citep{xie2026vadd}. Images are padded to $32\times32$, with each pixel represented as a binary token under masked diffusion. All three models use comparable UNet \citep{song2021sde} backbones and the same training budget. We report bits-per-dimension (BPD), the average negative log-likelihood per dimension in bits. Full implementation and training details are provided in Appendix~\ref{app:image-details}.

\begin{wraptable}{r}{0.35\linewidth}
  \centering
  \vspace{-1.8\baselineskip}
  \small
  \caption{Test Bits-per-dimension~($\downarrow$) and total parameter count on Binarized-MNIST.}
  \label{tab:mnist-bpd}
  \vspace{3pt}
  \begin{tabular}{lcc}
    \toprule
    Model & BPD~$\downarrow$ & Params \\
    \midrule
    MDLM & 0.077 & 2.07M \\
    VADD & 0.064 & 2.50M \\
    E-MoE (Ours) & \textbf{0.062} & 2.49M \\
    \bottomrule
  \end{tabular}
\end{wraptable}

\textbf{Results.} At low NFE, factorized MDLM produces fragmented strokes and globally inconsistent digit shapes, whereas both VADD and E-MoE already generate coherent samples. This mirrors the same factorization effect observed on the toy distributions, now in a much higher-dimensional discrete space. Table~\ref{tab:mnist-bpd} further shows that E-MoE achieves the best test BPD among the three methods while essentially matching VADD in parameter count. Full generation results across sampling steps are shown in Appendix~\ref{app:image-results}.

\subsection{Text generation}
\label{sec:lm1b}
\vspace{-2mm}
We now turn to unconditional text generation, following common practice for diffusion language models. We adopt the DUO setup~\citep{sahoo2025the} and train on LM1B~\citep{chelba2014lm1b} with sequence length $128$ for $1$M steps at global batch $512$, so every model sees $65$B tokens. All models share the same \texttt{small} DiT backbone, and E-MoE uses as many active parameters per token as MDLM. We compare the factorized MDLM~\citep{sahoo2024simple}, SEDD~\citep{lou2024sedd} and
BD3-LM~\citep{arriola2025bd3lm} (trained with different block sizes), VADD~\citep{xie2026vadd} with a continuous Gaussian latent, and E-MoE with a discrete shared latent. \underline{Full implementation LM1B text generation details} are given in Appendix~\ref{app:text-details}.

\begin{table}[t]
\centering
\footnotesize
\setlength{\tabcolsep}{3pt}
\renewcommand{\arraystretch}{1.1}
\begin{tabular*}{\textwidth}{@{\extracolsep{\fill}}c cc cc cc cc >{\columncolor{emoeblue!8}}c>{\columncolor{emoeblue!8}}c@{}}
\toprule
& \multicolumn{2}{c}{MDLM} & \multicolumn{2}{c}{SEDD} & \multicolumn{2}{c}{VADD} & \multicolumn{2}{c}{BD3-LM}
& \multicolumn{2}{c}{\textbf{E-MoE (ours)}}\\
\cmidrule(lr){2-3}\cmidrule(lr){4-5}\cmidrule(lr){6-7}\cmidrule(lr){8-9}\cmidrule(lr){10-11}
NFE & Gen-PPL$\downarrow$ & $H\uparrow$ & Gen-PPL$\downarrow$ & $H\uparrow$ & Gen-PPL$\downarrow$ & $H\uparrow$
& Gen-PPL$\downarrow$ & $H\uparrow$ & Gen-PPL$\downarrow$ & $H\uparrow$\\
\midrule
1   & 1433.8 & 4.37 & 1578.8 & 4.37 & 1270.8 & 4.34 & ---    & ---  & \textbf{643.8} & 4.35\\
2   &  997.5 & 4.37 &  1059.3 & 4.37 &  763.2 & 4.33 & ---    & ---  & \textbf{383.7} & 4.35\\
4   &  477.8 & 4.36 &  456.0 & 4.34 &  370.5 & 4.33 & ---    & ---  & \textbf{235.4} & 4.34\\
8   &  260.5 & 4.35 &  243.4 & 4.33 &  220.7 & 4.33 & 1130.6$^{16}$ & 4.34 & \textbf{174.9} & 4.34\\
16  &  179.4 & 4.35 &  166.7 & 4.33 &  168.4 & 4.33 & 1004.1$^{8}$ & 4.32 & \textbf{146.4} & 4.34\\
32  &  148.6 & 4.35 &  136.1 & 4.33 &  139.6 & 4.32 &  790.7$^{4}$ & 4.29 & \textbf{135.1} & 4.34\\
64  &  134.1 & 4.35 & \textbf{125.6} & 4.32 &  127.4 & 4.32 & ---    & ---  & 128.1 & 4.33\\
128 &  126.3 & 4.35 & 118.5 & 4.32 & \textbf{118.4} & 4.32 & ---    & ---  & 126.3 & 4.34\\
\midrule
AR  & \multicolumn{10}{c}{$67.97$\quad($H=4.32$) at NFE $=128$}\\
\bottomrule
\end{tabular*}
\vspace{-3mm}
\caption{\textbf{Few-step generation on LM1B.} Generative perplexity (Gen-PPL) and sample entropy $H$
(data: $4.32$), averaged over $2000$ samples with categorical sampling in \texttt{fp64}. Superscripts
mark BD3-LM models trained with different block sizes. Best per row in \textbf{bold}.}
\label{tab:lm1b-genppl-main}
\end{table}
\begin{figure}[t]
  \centering
  \begin{minipage}[b]{0.5\linewidth}
    \centering
    \includegraphics[width=\linewidth]{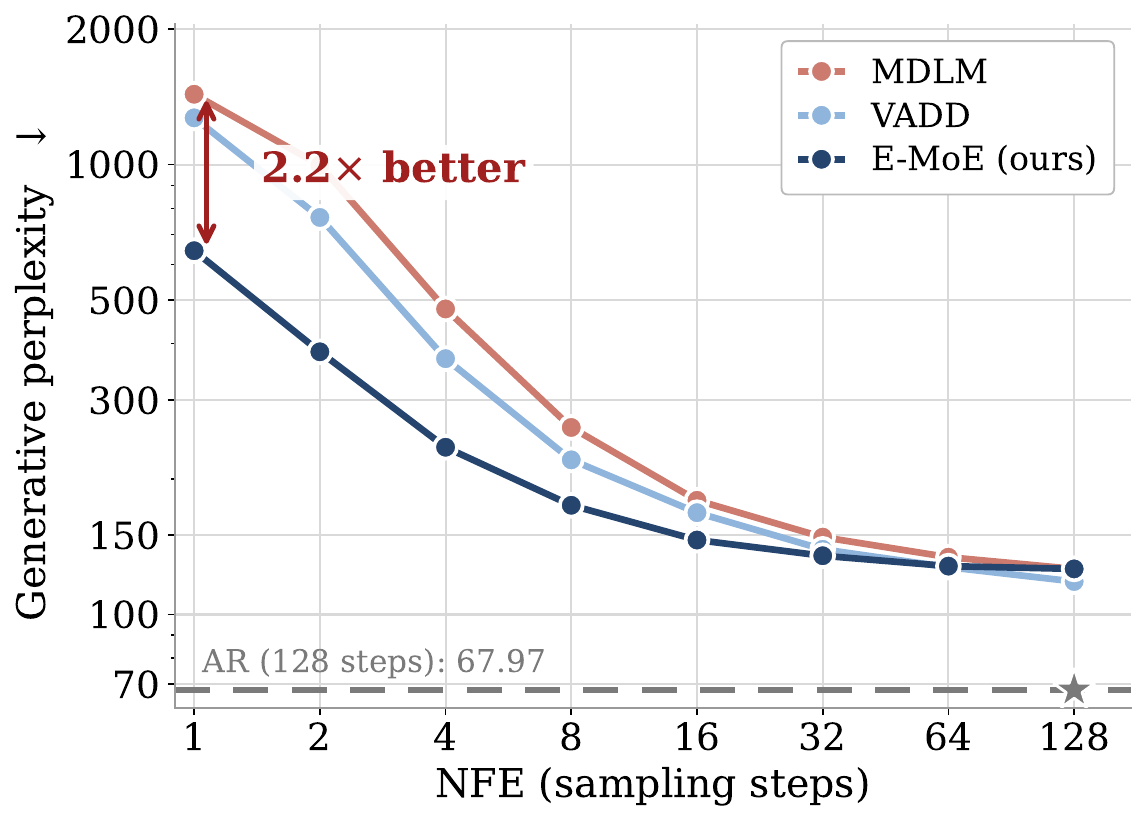}
  \end{minipage}\hfill
  \begin{minipage}[b]{0.5\linewidth}
    \centering
    \includegraphics[width=\linewidth]{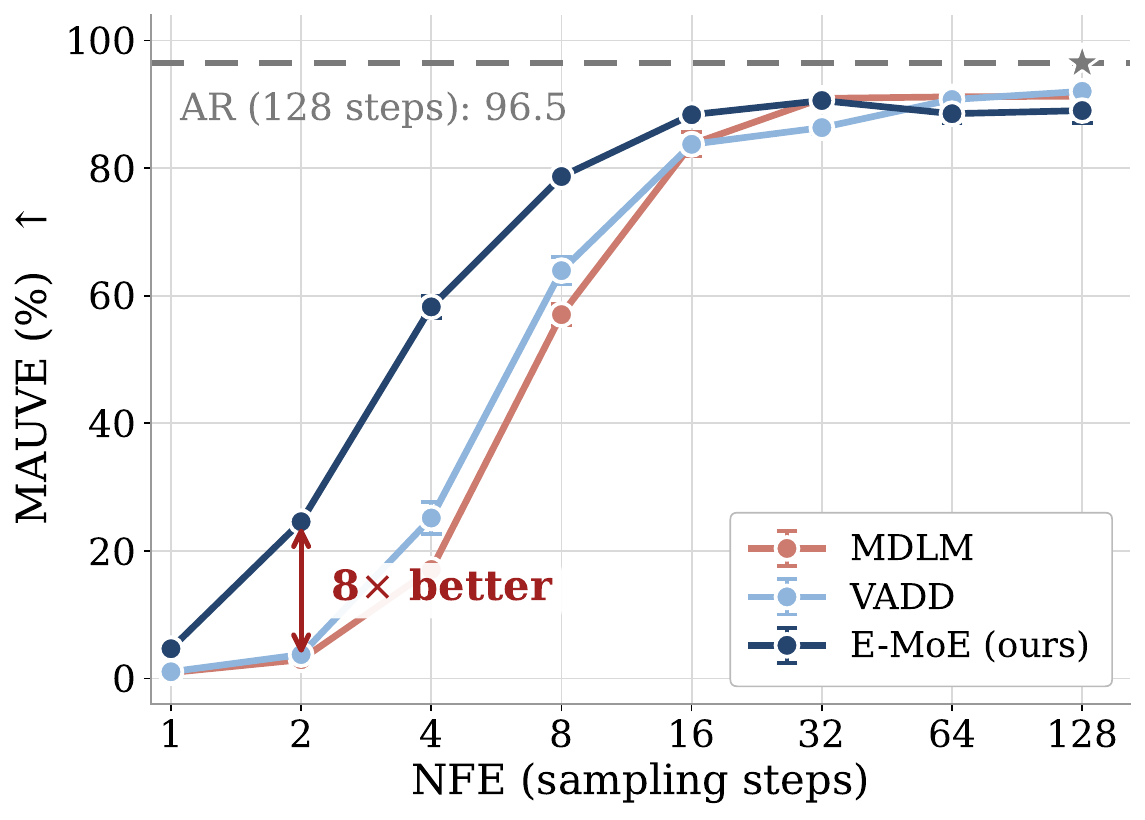}
  \end{minipage}
  \vspace{-2mm}
  \caption{\textbf{Two views of the same LM1B sweep.} \emph{Left:} Generative Perplexity ($\downarrow$). \emph{Right:} MAUVE ($\uparrow$)}
  \label{fig:lm1b-pair}
\end{figure}

\vspace{-2mm}
\paragraph{Results.} The gain is largest where the factorization barrier binds (Table~\ref{tab:lm1b-genppl-main}, Figure~\ref{fig:lm1b-pair}). At NFE~$=1$ and $2$, E-MoE lowers generative perplexity by $2$--$2.6\times$ relative to every baseline, at the same sample entropy ($4.34$--$4.35$). MAUVE separates the models even more: at NFE~$=2$ E-MoE reaches $24.6\%$ against $3.0\%$ for MDLM and $3.8\%$ for VADD. The advantage shrinks as NFE grows, as expected once each step unmasks few tokens. \textbf{A discrete shared latent makes few-step generation markedly better at no extra cost in active parameters.} \underline{Ablation, additional results, and generated text-samples} are given in Appendix~\ref{app:text-results}.

\vspace{-4mm}
\section{Discussion}
\label{sec:discussion}
\vspace{-2mm}
We develop an MDM based on a mixture of factorized marginals conditioned on discrete latent codes drawn from an informative data prior, which is learned with no extra models, using the same network as for clean token prediction, in order to model correlations between predicted tokens. Marginalizing over routes couples tokens unmasked in the same step, which helps break the factorization barrier: E-MoE improves few-step generation on synthetic densities, binarized MNIST and LM1B, lowering generative perplexity on LM1B by at least $2\times$ at one and two steps.

\section*{AI Use Statement}
AI tools assisted with polishing the text and checking  proofs. All AI-assisted work was reviewed by the authors, who take the responsibility for the final content for this work.

\section*{Ethics Statement}
This work focuses on methodological developments for discrete diffusion language models and does not involve human subjects, personally identifiable information, or sensitive data. The experiments use synthetic benchmarks and publicly available
datasets. We are not aware of any specific ethical risks beyond those associated with the general use of machine learning methods, and we have aimed to report the methodology, experimental setup, and results transparently.

\section*{Reproducibility Statement}
To ensure reproducibility, We provide the experimental details in Appendices B, C and the code to reproduce the conducted experiments in the supplementary materials.

\bibliographystyle{iclr2027_conference}
\bibliography{references}

\newpage
\appendix

\section{Theoretical results and proofs}
\label{app:derivations}

Appendix~\ref{app:mdm-prelim} derives the masked-diffusion ELBO of
Section~\ref{sec:background}. Appendices~\ref{app:prop1} and~\ref{app:prop2}
prove Propositions~1 and~2, which extend it to E-MoE's mixture
parameterization and give the objective \eqref{eq:emoe-objective}.

\subsection{ELBO for masked diffusion models}
\label{app:mdm-prelim}

We derive \eqref{eq:mdlm_fwd}, \eqref{eq:mdlm_back_posterior}, \eqref{eq:factor}
and the resulting ELBO from the absorbing kernel. Everything factorizes over positions, so we fix $\ell$ and drop it where unambiguous. We write $\langle\pi,\textbf{v}\rangle$ for the mass that $\mathrm{Cat}(\cdot;\pi)$ puts on a one-hot $\textbf{v}$, and take $\alpha_t$ strictly decreasing with $\alpha_0=1$, $\alpha_1=0$.

\subsubsection{Forward marginals}
\label{app:forward}

The kernel of \eqref{eq:mdlm_fwd} is
$q(\textbf{x}_t^\ell\mid\textbf{x}_s^\ell)
=\mathrm{Cat}(\textbf{x}_t^\ell;\alpha_{t|s}\textbf{x}_s^\ell+(1-\alpha_{t|s})\textbf{m})$.
Setting $\textbf{x}_s^\ell=\textbf{m}$ gives
$\alpha_{t|s}\textbf{m}+(1-\alpha_{t|s})\textbf{m}=\textbf{m}$, so $\textbf{m}$ is
absorbing. Marginalizing the intermediate state,
\begin{align}
  q(\textbf{x}_t^\ell\mid\textbf{x}_0^\ell)
  &=
  \sum_{\textbf{x}_s^\ell}
  q(\textbf{x}_t^\ell\mid\textbf{x}_s^\ell)\,
  q(\textbf{x}_s^\ell\mid\textbf{x}_0^\ell)
  =
  \alpha_s\left[\alpha_{t|s}\textbf{x}_0^\ell+(1-\alpha_{t|s})\textbf{m}\right]
  +(1-\alpha_s)\,\textbf{m}
  \nonumber\\
  &=
  \alpha_s\alpha_{t|s}\,\textbf{x}_0^\ell
  +\bigl[\alpha_s(1-\alpha_{t|s})+1-\alpha_s\bigr]\textbf{m}
  =
  \mathrm{Cat}\!\left(\textbf{x}_t^\ell;\,
  \alpha_t\textbf{x}_0^\ell+(1-\alpha_t)\textbf{m}\right),
  \label{eq:app-marginal}
\end{align}
where the first line uses $\mathrm{supp}\,q(\textbf{x}_s^\ell\mid\textbf{x}_0^\ell)
=\{\textbf{x}_0^\ell,\textbf{m}\}$ and the absorbing property, and the last step
uses $\alpha_s\alpha_{t|s}=\alpha_t$. Hence each position is masked independently
with probability $1-\alpha_t$, and $\alpha_1=0$ makes
$q(\textbf{x}_1\mid\textbf{x}_0)$ the point mass on the fully masked sequence.

\subsubsection{Reverse posterior}
\label{app:posterior}

By Bayes' rule,
$q(\textbf{x}_s^\ell\mid\textbf{x}_t^\ell,\textbf{x}_0^\ell)
= q(\textbf{x}_t^\ell\mid\textbf{x}_s^\ell)\,
  q(\textbf{x}_s^\ell\mid\textbf{x}_0^\ell)\,/\,
  q(\textbf{x}_t^\ell\mid\textbf{x}_0^\ell)$.
If $\textbf{x}_t^\ell\neq\textbf{m}$ then $\textbf{x}_t^\ell=\textbf{x}_0^\ell$,
and the numerator vanishes unless $\textbf{x}_s^\ell=\textbf{x}_t^\ell$, since
$\textbf{m}$ is absorbing:
\begin{equation}
  q(\textbf{x}_s^\ell\mid\textbf{x}_t^\ell\neq\textbf{m},\textbf{x}_0^\ell)
  =\mathrm{Cat}(\textbf{x}_s^\ell;\textbf{x}_t^\ell).
  \label{eq:app-post-unmasked}
\end{equation}
If $\textbf{x}_t^\ell=\textbf{m}$ then
$\textbf{x}_s^\ell\in\{\textbf{x}_0^\ell,\textbf{m}\}$ and, with
\eqref{eq:app-marginal} in the denominator,
\begin{equation}
  q(\textbf{x}_s^\ell=\textbf{x}_0^\ell\mid\textbf{x}_t^\ell=\textbf{m},\textbf{x}_0^\ell)
  =\frac{(1-\alpha_{t|s})\,\alpha_s}{1-\alpha_t}
  =\frac{\alpha_s-\alpha_t}{1-\alpha_t},
  \qquad
  q(\textbf{x}_s^\ell=\textbf{m}\mid\textbf{x}_t^\ell=\textbf{m},\textbf{x}_0^\ell)
  =\frac{1-\alpha_s}{1-\alpha_t},
  \label{eq:app-post-masked}
\end{equation}
which sum to $1$. Together \eqref{eq:app-post-unmasked} and
\eqref{eq:app-post-masked} are \eqref{eq:mdlm_back_posterior}.

\subsubsection{Reverse model}
\label{app:subs}

Replacing $\textbf{x}_0^\ell$ in \eqref{eq:mdlm_back_posterior} by a prediction
$\textbf{x}_\theta^\ell(\textbf{x}_t,t)\in\triangle^{K}$ computed from the whole
noisy sequence gives \eqref{eq:factor}. Following \citet{sahoo2024simple} we
impose
\begin{equation}
  \langle\textbf{x}_\theta^\ell(\textbf{x}_t,t),\textbf{m}\rangle=0,
  \qquad
  \textbf{x}_\theta^\ell(\textbf{x}_t,t)=\textbf{x}_t^\ell
  \ \ \text{if}\ \ \textbf{x}_t^\ell\neq\textbf{m}.
  \label{eq:app-subs}
\end{equation}
The second condition makes the $\ell$-th factor of \eqref{eq:factor} equal
\eqref{eq:app-post-unmasked} at unmasked positions; the first makes it put mass
$(1-\alpha_s)/(1-\alpha_t)$ on $\textbf{m}$ at masked positions, matching
\eqref{eq:app-post-masked}. Only the event that a masked position is revealed can
therefore contribute to a KL between $q$ and $p_\theta$.

\subsubsection{Discrete-time ELBO}
\label{app:nelbo-discrete}

Fix a grid $0=t_0<\dots<t_T=1$ and write $s=t_{i-1}$, $t=t_i$,
$a_i:=\dfrac{\alpha_{t_{i-1}}-\alpha_{t_i}}{1-\alpha_{t_i}}$.
\begin{equation}
\begin{aligned}
  -\log p_\theta(\textbf{x}_0)
  \;\le\;
  &\underbrace{\mathbb{E}_{q}\!\left[-\log p_\theta(\textbf{x}_0\mid\textbf{x}_{t_1})\right]}
  _{\textstyle\mathcal{L}_{\mathrm{recon}}}
  \;+\;
  \underbrace{\mathrm{KL}\!\left(q(\textbf{x}_{t_T}\mid\textbf{x}_0)
  \,\|\,p_\theta(\textbf{x}_{t_T})\right)}_{\textstyle\mathcal{L}_{\mathrm{prior}}\;=\;0}
  \\[6pt]
  +\;
  &\underbrace{\sum_{i=2}^{T}\mathbb{E}_{q}\,
  \mathrm{KL}\!\left(q(\textbf{x}_{t_{i-1}}\mid\textbf{x}_{t_i},\textbf{x}_0)
  \,\|\,p_\theta(\textbf{x}_{t_{i-1}}\mid\textbf{x}_{t_i})\right)}
  _{\textstyle\mathcal{L}_{\mathrm{diff}}}
  \qquad\text{(Appendix~\ref{app:prop1})},
\end{aligned}
  \label{eq:app-nelbo-split}
\end{equation}
where $\mathcal{L}_{\mathrm{prior}}=0$ because $\alpha_1=0$ makes both arguments
the fully masked point mass. The per-step KL evaluates to
\begin{alignat}{2}
  &\mathrm{KL}\!\left(q(\textbf{x}_s\mid\textbf{x}_t,\textbf{x}_0)
  \,\|\,p_\theta(\textbf{x}_s\mid\textbf{x}_t)\right)
  \nonumber\\[2pt]
  &\quad=\;
  \sum_{\ell=1}^{L}
  \mathrm{KL}\!\left(q(\textbf{x}_s^\ell\mid\textbf{x}_t,\textbf{x}_0)
  \,\|\,p_\theta(\textbf{x}_s^\ell\mid\textbf{x}_t)\right)
  &&\text{\small both sides factorize over $\ell$}
  \nonumber\\[2pt]
  &\quad=\;
  \sum_{\ell\in\mathcal{M}(\textbf{x}_t)}
  \mathrm{KL}\!\left(q(\textbf{x}_s^\ell\mid\textbf{x}_t,\textbf{x}_0)
  \,\|\,p_\theta(\textbf{x}_s^\ell\mid\textbf{x}_t)\right)
  &&\text{\small $\textbf{x}_\theta^\ell=\textbf{x}_t^\ell$ on $\ell\notin\mathcal{M}$ by \eqref{eq:app-subs}}
  \nonumber\\[2pt]
  &\quad=\;
  \sum_{\ell\in\mathcal{M}(\textbf{x}_t)}
  \left[
  a_i\log\frac{a_i}{a_i\langle\textbf{x}_\theta^\ell,\textbf{x}_0^\ell\rangle}
  +(1-a_i)\log\frac{1-a_i}{1-a_i}
  \right]
  \quad
  &&\text{\small by \eqref{eq:app-post-masked} and $\langle\textbf{x}_\theta^\ell,\textbf{m}\rangle=0$}
  \nonumber\\[2pt]
  &\quad=\;
  -\,a_i\!\!\sum_{\ell\in\mathcal{M}(\textbf{x}_t)}\!\!
  \log\left\langle\textbf{x}_\theta^\ell,\textbf{x}_0^\ell\right\rangle .
  &&
  \label{eq:app-kl-pertoken}
\end{alignat}
Since $\alpha_{t_0}=1$, the $i=1$ term of \eqref{eq:app-nelbo-split} equals
\eqref{eq:app-kl-pertoken} with $a_1=1$, so
\begin{equation}
  \mathcal{L}_{T}(\textbf{x}_0,\theta)
  =
  \sum_{i=1}^{T}
  \mathbb{E}_{q(\textbf{x}_{t_i}\mid\textbf{x}_0)}
  \left[
  \frac{\alpha_{t_{i-1}}-\alpha_{t_i}}{1-\alpha_{t_i}}
  \sum_{\ell\in\mathcal{M}(\textbf{x}_{t_i})}
  -\log\left\langle\textbf{x}_\theta^\ell(\textbf{x}_{t_i},t_i),
  \textbf{x}_0^\ell\right\rangle
  \right].
  \label{eq:app-nelbo-discrete}
\end{equation}

\subsubsection{Continuous-time limit}
\label{app:nelbo-continuous}

Abbreviate the inner sum of \eqref{eq:app-nelbo-discrete} as
\begin{equation}
  G_\theta(\textbf{x}_t,t)
  :=
  \sum_{\ell\in\mathcal{M}(\textbf{x}_t)}
  \log\left\langle\textbf{x}_\theta^\ell(\textbf{x}_t,t),\textbf{x}_0^\ell\right\rangle
  \;\le\;0,
  \qquad\text{so}\qquad
  \mathcal{L}_{T}
  =
  -\sum_{i=1}^{T}
  \frac{\alpha_{t_{i-1}}-\alpha_{t_i}}{1-\alpha_{t_i}}\,
  \mathbb{E}_{q}\!\left[G_\theta(\textbf{x}_{t_i},t_i)\right].
  \label{eq:app-G}
\end{equation}
Take $t_i=i/T$ and let $T\to\infty$:
\begin{alignat}{2}
  \alpha_{t_{i-1}}-\alpha_{t_i}
  &\;=\; -\frac{\alpha'_{t_i}}{T}+o\!\left(\tfrac{1}{T}\right)
  &&\qquad\text{\small ($\alpha_t$ differentiable)}
  \label{eq:app-taylor}\\[4pt]
  \mathcal{L}_{T}
  &\;=\;
  \sum_{i=1}^{T}\frac{1}{T}\,
  \frac{\alpha'_{t_i}}{1-\alpha_{t_i}}\,
  \mathbb{E}_{q}\!\left[G_\theta(\textbf{x}_{t_i},t_i)\right]+o(1)
  &&\qquad\text{\small (into \eqref{eq:app-G})}
  \label{eq:app-riemann}\\[4pt]
  \mathcal{L}_{\infty}
  &\;=\;
  \int_{0}^{1}
  \frac{\alpha'_t}{1-\alpha_t}\,
  \mathbb{E}_{q(\textbf{x}_t\mid\textbf{x}_0)}
  \!\left[G_\theta(\textbf{x}_t,t)\right]dt
  &&\qquad\text{\small (Riemann sum)}
  \label{eq:app-nelbo-continuous}
\end{alignat}
with $\mathcal{L}_\infty\ge0$ since $\alpha'_t<0$ and $G_\theta\le0$. Expanding
$G_\theta$ recovers the ELBO term of $\mathcal{L}_{\text{vadd}}$ in
Section~\ref{sec:background} with $\textbf{z}$ removed. The product form entered
only once, in the second line of \eqref{eq:app-kl-pertoken};
Appendix~\ref{app:prop2} repeats that step for a reverse model that is a mixture
of such products.

\subsection{Proof of Lemma 1}
\label{app:prop1}

\begin{findingbox}{Lemma 1: Upper bound for NLL}
For any discretization $t_0<\cdots<t_n$ with
$\textbf{x}_{t_0}=\textbf{x}$,
\begin{equation}
\mathbb{E}_{p_{data}(\textbf{x}_{0})}[-\log p_\theta(\textbf{x}_{0})] \;\le\; \mathbb{E}_{p_{data}(\textbf{x}_{0})} \sum_{i=1}^{T}  \mathrm{KL}\!\left(q(\textbf{x}_{t_{i-1}} \mid \textbf{x}_{t_i}, \textbf{x}_{0}) \,\middle\|\, p_\theta(\textbf{x}_{t_{i-1}} \mid \textbf{x}_{t_i})\right).
\label{eq:diffusion-elbo}
\end{equation}
\end{findingbox}

\textbf{Proof.}

1)  We derive an upper bound for negative log-likelihood $-\log p_{\theta}(\textbf{x}_{0})$, using the following:

$-\log p_{\theta}(\textbf{x}_{0}) = -\log \int_{\textbf{z}} p_{\theta}(\textbf{x}_{0},\textbf{z} )d\textbf{z} = -\log \int_{\textbf{z}} p_{\theta}(\textbf{x}_{0},\textbf{z} ) \frac{q(\textbf{z}|\textbf{x}_{0})}{q(\textbf{z}|\textbf{x}_{0})}d\textbf{z} = - \mathbb{E}_{q(\textbf{z}|\textbf{x}_{0})}\log\frac{ p_{\theta}(\textbf{x}_{0},\textbf{z} ) }{q(\textbf{z}|\textbf{x}_{0})} $

Then, using Jensen inequality: $-\log p_{\theta}(\textbf{x}_{0}) = - \mathbb{E}_{q(\textbf{z}|\textbf{x}_{0})}\log\frac{ p_{\theta}(\textbf{x}_{0},\textbf{z} ) }{q(\textbf{z}|\textbf{x}_{0})} \leq - \log\mathbb{E}_{q(\textbf{z}|\textbf{x}_{0})}\frac{ p_{\theta}(\textbf{x}_{0},\textbf{z} ) }{q(\textbf{z}|\textbf{x}_{0})}$

2) Since we consider diffusion process, then latent $\textbf{z}$ is a set of intermediate values between data and full-mask state as $\{\textbf{x}_{t_1},...,\textbf{x}_{t_T} \}$, where $\textbf{x}_{t_0} =  \textbf{x}_{0}$. Then, rewrite the last inequality via this notation:

$$ -\log p_{\theta}(\textbf{x}_{0}) \leq - \int q(\textbf{x}_{t_1},...,\textbf{x}_{t_T}|\textbf{x}_{0})[\log p_{\theta}(\textbf{x}_{0},\textbf{x}_{t_1},...,\textbf{x}_{t_T})  - \log q(\textbf{x}_{t_1},...,\textbf{x}_{t_T}|\textbf{x}_{0}) ]d\textbf{x}_{t_1}...d\textbf{x}_{t_T} $$

Since the diffusion process is Markovian, then we rewrite the last expression using this property:

$$ -\log p_{\theta}(\textbf{x}_{0}) \leq - \int  \prod_{i=1}^{T}q(\textbf{x}_{t_i}|\textbf{x}_{t_{i-1}})[\log p_{\theta}(\textbf{x}_{t_T})  + \sum_{i=1}^{T}\log p_{\theta}(\textbf{x}_{t_{i-1}}|\textbf{x}_{t_{i}}) - \sum_{i=1}^{T}\log q(\textbf{x}_{t_i}|\textbf{x}_{t_{i-1}}) ]d\textbf{x}_{t_1}...d\textbf{x}_{t_T} $$

3) Rewriting $q(\textbf{x}_{t_i}\mid \textbf{x}_{t_{i-1}})$ via Bayes' rule in terms of $q(\textbf{x}_{t_i}\mid \textbf{x}_{0})$, $q(\textbf{x}_{t_{i-1}}\mid \textbf{x}_{t_i},\textbf{x}_{0})$, $q(\textbf{x}_{t_{i-1}}\mid \textbf{x}_{0})$ and assuming $q(\textbf{x}_{t_1}|\textbf{x}_{0}) \approx 1$ and telescoping the resulting ratio over $i$ collapses this to
\begin{equation}
-\log p_\theta(\textbf{x}_{0}) \;\le\; -\mathbb{E}_{q(\textbf{x}_{t_1},...,\textbf{x}_{t_T}\mid \textbf{x}_{0})}\Big[\log\frac{p(\textbf{x}_{t_T})}{q(\textbf{x}_{t_T}\mid \textbf{x}_{0})} + \sum_{i=1}^T \log\frac{p_\theta(\textbf{x}_{t_{i-1}}\mid \textbf{x}_{t_i})}{q(\textbf{x}_{t_{i-1}}\mid \textbf{x}_{t_i},\textbf{x}_{0})})\Big].
\end{equation}

Since the first term is $\mathrm{KL}(q(\textbf{x}_{t_T}\mid \textbf{x}_{0})\|p(\textbf{x}_{t_T}))=0$ (both sides equal the fully-masked prior), so:

$$ -\log p_{\theta}(\textbf{x}_{0}) \leq  \sum_{i=1}^{T} \text{KL} \left(q(\textbf{x}_{t_{i-1}} \mid \textbf{x}_{t_i}, \textbf{x}_{0}) \,\middle\|\, p_\theta(\textbf{x}_{t_{i-1}} \mid \textbf{x}_{t_i})\right)$$

Applying the expectation over $p_{data}(\textbf{x}_{0})$ we get the statement:

$$ \mathbb{E}_{p_{data}(\textbf{x}_{0})}[-\log p_{\theta}(\textbf{x}_{0})] \leq  \mathbb{E}_{p_{data}(\textbf{x}_{0})} \sum_{i=1}^{T} \text{KL} \left(q(\textbf{x}_{t_{i-1}} \mid \textbf{x}_{t_i}, \textbf{x}_{0}) \,\middle\|\, p_\theta(\textbf{x}_{t_{i-1}} \mid \textbf{x}_{t_i})\right).$$

\subsection{Proof of Proposition 3.1}
\label{app:prop2}

\begin{findingbox}{Proposition 3.1}
\label{text_prop} Let $i \in \overline{1,T}$ is a number of diffusion step and $t_i:= \frac{i}{T}$ is a  $i$-th time step. Let $p_{\text{data}}(\textbf{x}_{0})$ is a data distribution of $L$-length sequences and $p_{\theta}(\textbf{x}_{0})$ is its parametric estimation. Let $\textbf{x}_{\theta} \sim \mu_{\theta}(\textbf{x}_{t_i}, t_{i})$ is a estimation of clean $\textbf{x}_{0}$ and $\mathcal{M}(\textbf{x}_t)$ is a masked subset of $\textbf{x}_t$. Let $p_{\theta}(\textbf{z}|\textbf{x}_{t_i}, \textbf{x}_{0})$ and $q_{\theta} (\textbf{z}|\textbf{x}_{t_i})$ are  posterior and prior distributions at $i$-th step. \\

Then, the upper bound of negative log-likelihhod $\mathbb{E}_{p_{\text{data}(\textbf{x}_{0})}}[ - \log p_{\theta}(\textbf{x}_{0})]$ is given by:

\begin{equation}
\label{eq:our_loss}
\mathbb{E}_{p_{data}(\textbf{x}_{0})} \frac{1}{\text{T}} \sum_{i=1}^{\text{T}} \mathbb{E}_{q(\textbf{x}_{t_i}|\textbf{x}_{0})}\mathbb{E}_{q_{\theta}(\textbf{z}|\textbf{x}_{t_i},\textbf{x}_{0})} [\frac{-1}{t_i}\sum_{\ell \in \mathcal{M}(\textbf{x}_t)} \log p_{\theta}(\textbf{x}_{0}^{\ell}|\textbf{x}^{\ell}_{t_i},\textbf{z}) + \text{T}\log\frac{q_{\theta}(\textbf{z}|\textbf{x}_{t_i},\textbf{x}_{0})}{p_{\theta}(\textbf{z}|\textbf{x}_{t_i})}]
 \end{equation}
\end{findingbox}

\textbf{Proof.}

Lemma~1 has already shown that the negative log-likelihood is upper bounded by
the sum of per-step KL divergences
\[
\mathrm{KL}\!\left(
q(\textbf{x}_{t_{i-1}}\mid\textbf{x}_{t_i},\textbf{x}_{0})
\,\middle\|\,
p_{\theta}(\textbf{x}_{t_{i-1}}\mid\textbf{x}_{t_i})
\right).
\]
Thus, it remains to upper bound this KL for the mixture reverse model and then
substitute the result back into Lemma~1.

1) We first derive a lower bound on
$\log p_{\theta}(\textbf{x}_{t_{i-1}}\mid\textbf{x}_{t_i})$.
For readability, fix a time step $i$ and introduce the auxiliary route
distribution
\[
  r_i(\textbf{z})
  :=
  q_{\theta}(\textbf{z}\mid\textbf{x}_{t_i},\textbf{x}_{0}).
\]
The generative route distribution used by the prior path is
$p_{\theta}(\textbf{z}\mid\textbf{x}_{t_i},\textbf{x}_{\theta})$.
Using Jensen's inequality,
\begin{equation}
\begin{aligned}
\log p_{\theta}(\textbf{x}_{t_{i-1}}\mid\textbf{x}_{t_i})
&=
\log
\sum_{\textbf{z}}
p_{\theta}(\textbf{x}_{t_{i-1}},\textbf{z}\mid\textbf{x}_{t_i})
\\
&=
\log
\sum_{\textbf{z}}
r_i(\textbf{z})
\frac{
p_{\theta}(\textbf{x}_{t_{i-1}},\textbf{z}\mid\textbf{x}_{t_i})
}{
r_i(\textbf{z})
}
\\
&\geq
\mathbb{E}_{r_i(\textbf{z})}
\left[
\log
\frac{
p_{\theta}(\textbf{x}_{t_{i-1}},\textbf{z}\mid\textbf{x}_{t_i})
}{
r_i(\textbf{z})
}
\right].
\end{aligned}
\label{eq:latent-lower-bound}
\end{equation}
Now we expand the joint term as
\begin{equation}
\begin{aligned}
p_{\theta}(\textbf{x}_{t_{i-1}},\textbf{z}\mid\textbf{x}_{t_i})
&=
p_{\theta}(\textbf{x}_{t_{i-1}}\mid\textbf{x}_{t_i},\textbf{z})\,
p_{\theta}(\textbf{z}\mid\textbf{x}_{t_i}),
\\
\log
\frac{
p_{\theta}(\textbf{x}_{t_{i-1}},\textbf{z}\mid\textbf{x}_{t_i})
}{
r_i(\textbf{z})
}
&=
\log p_{\theta}(\textbf{x}_{t_{i-1}}\mid\textbf{x}_{t_i},\textbf{z})
\;+\;
\log
\frac{
p_{\theta}(\textbf{z}\mid\textbf{x}_{t_i} )
}{
q_{\theta}(\textbf{z}\mid\textbf{x}_{t_i},\textbf{x}_{0})
}.
\end{aligned}
\label{eq:joint-factor}
\end{equation}
Substituting \eqref{eq:joint-factor} into \eqref{eq:latent-lower-bound} gives
\begin{equation}
\begin{aligned}
\log p_{\theta}(\textbf{x}_{t_{i-1}}\mid\textbf{x}_{t_i})
\geq\;&
\mathbb{E}_{q_{\theta}(\textbf{z}\mid\textbf{x}_{t_i},\textbf{x}_{0})}
\left[
\log p_{\theta}(\textbf{x}_{t_{i-1}}\mid\textbf{x}_{t_i},\textbf{z})
\right]
\\
&-
\mathrm{KL}\!\left(
q_{\theta}(\textbf{z}\mid\textbf{x}_{t_i},\textbf{x}_{0})
\,\middle\|\,
p_{\theta}(\textbf{z}\mid\textbf{x}_{t_i})
\right).
\end{aligned}
\label{eq:latent-elbo}
\end{equation}
Equivalently,
\begin{equation}
\begin{aligned}
-\log p_{\theta}(\textbf{x}_{t_{i-1}}\mid\textbf{x}_{t_i})
\leq\;&
\mathrm{KL}\!\left(
    q_{\theta}(\textbf{z}\mid\textbf{x}_{t_i},\textbf{x}_{0})
\,\middle\|\,
p_{\theta}(\textbf{z}\mid\textbf{x}_{t_i},\textbf{x}_{\theta})
\right)
\\
&-
\mathbb{E}_{q_{\theta}(\textbf{z}\mid\textbf{x}_{t_i},\textbf{x}_{0})}
\left[
\log p_{\theta}(\textbf{x}_{t_{i-1}}\mid\textbf{x}_{t_i},\textbf{z})
\right].
\end{aligned}
\label{eq:latent-neg-bound}
\end{equation}

2) We now apply this inequality inside the per-step KL from Lemma~1:
\begin{equation}
\begin{aligned}
&\mathrm{KL}\!\left(
q(\textbf{x}_{t_{i-1}}\mid\textbf{x}_{t_i},\textbf{x}_{0})
\,\middle\|\,
p_{\theta}(\textbf{x}_{t_{i-1}}\mid\textbf{x}_{t_i})
\right)
\\
&\quad =
\mathbb{E}_{q(\textbf{x}_{t_{i-1}}\mid\textbf{x}_{t_i},\textbf{x}_{0})}
\left[
\log q(\textbf{x}_{t_{i-1}}\mid\textbf{x}_{t_i},\textbf{x}_{0})
-\log p_{\theta}(\textbf{x}_{t_{i-1}}\mid\textbf{x}_{t_i})
\right].
\end{aligned}
\label{eq:kl-target}
\end{equation}
Since $f\leq g$ implies $\mathbb{E}f\leq\mathbb{E}g$, substituting
\eqref{eq:latent-neg-bound} into \eqref{eq:kl-target} gives
\begin{equation}
\begin{aligned}
&\mathrm{KL}\!\left(
q(\textbf{x}_{t_{i-1}}\mid\textbf{x}_{t_i},\textbf{x}_{0})
\,\middle\|\,
p_{\theta}(\textbf{x}_{t_{i-1}}\mid\textbf{x}_{t_i})
\right)
\\
&\leq
\mathbb{E}_{p_{\theta}(\textbf{z}\mid\textbf{x}_{t_i},\textbf{x}_{0})}
\left[
\mathrm{KL}\!\left(
q(\textbf{x}_{t_{i-1}}\mid\textbf{x}_{t_i},\textbf{x}_{0})
\,\middle\|\,
p_{\theta}(\textbf{x}_{t_{i-1}}\mid\textbf{x}_{t_i},\textbf{z})
\right)
\right]
\\
&\qquad
+
\mathrm{KL}\!\left(
q_{\theta}(\textbf{z}\mid\textbf{x}_{t_i},\textbf{x}_{0})
\,\middle\|\,
p_{\theta}(\textbf{z}\mid\textbf{x}_{t_i})
\right).
\end{aligned}
\label{eq:kl-split}
\end{equation}
The route-KL term does not depend on
$\textbf{x}_{t_{i-1}}$ under
$q(\textbf{x}_{t_{i-1}}\mid\textbf{x}_{t_i},\textbf{x}_{0})$, so it comes out of
that expectation.

3) Next we substitute the mixture component. For a fixed route $\textbf{z}$, the
conditional reverse process factorizes over positions:
\begin{equation}
  p_{\theta}(\textbf{x}_{t_{i-1}}\mid\textbf{x}_{t_i},\textbf{z})
  =
  \prod_{\ell=1}^{L}
  p_{\theta}(\textbf{x}_{t_{i-1}}^{\ell}\mid
  \textbf{x}_{t_i},\textbf{z}).
\label{eq:component-factor}
\end{equation}
At the same time, the MDLM posterior also factorizes over positions. Therefore,
for a fixed $\textbf{z}$,
\begin{equation}
\begin{aligned}
&\mathrm{KL}\!\left(
q(\textbf{x}_{t_{i-1}}\mid\textbf{x}_{t_i},\textbf{x}_{0})
\,\middle\|\,
p_{\theta}(\textbf{x}_{t_{i-1}}\mid\textbf{x}_{t_i},\textbf{z})
\right)
\\
&\qquad =
\sum_{\ell=1}^{L}
\mathrm{KL}\!\left(
q(\textbf{x}_{t_{i-1}}^{\ell}\mid\textbf{x}_{t_i},\textbf{x}_{0})
\,\middle\|\,
p_{\theta}(\textbf{x}_{t_{i-1}}^{\ell}\mid\textbf{x}_{t_i},\textbf{z})
\right).
\end{aligned}
\label{eq:factor-kl}
\end{equation}
If $\ell\notin\mathcal{M}(\textbf{x}_{t_i})$, then
$\textbf{x}_{t_i}^{\ell}\neq\textbf{m}$ and the true posterior is the point mass
on $\textbf{x}_{t_i}^{\ell}$; the reverse model uses the same unmasked token, so
the corresponding KL is zero. Thus only masked positions contribute.

4) For $\ell\in\mathcal{M}(\textbf{x}_{t_i})$, the MDLM posterior from
\eqref{eq:mdlm_back_posterior} gives
\begin{equation}
\begin{aligned}
q(\textbf{x}_{t_{i-1}}^{\ell}=\textbf{x}_{0}^{\ell}
\mid\textbf{x}_{t_i},\textbf{x}_{0})
&=
\frac{\alpha_{t_{i-1}}-\alpha_{t_i}}{1-\alpha_{t_i}},
\\
q(\textbf{x}_{t_{i-1}}^{\ell}=\textbf{m}
\mid\textbf{x}_{t_i},\textbf{x}_{0})
&=
\frac{1-\alpha_{t_{i-1}}}{1-\alpha_{t_i}}.
\end{aligned}
\label{eq:posterior-two-masses}
\end{equation}
Conditioned on $\textbf{z}$, the E-MoE reverse component uses the same absorbing
form, but replaces the unknown clean token by the expert prediction. Hence the
probability assigned to the clean token is
\begin{equation}
  p_{\theta}(
  \textbf{x}_{t_{i-1}}^{\ell}=\textbf{x}_{0}^{\ell}
  \mid \textbf{x}_{t_i},\textbf{z})
  =
  \frac{\alpha_{t_{i-1}}-\alpha_{t_i}}{1-\alpha_{t_i}}\,
  p_{\theta}(\textbf{x}_{0}^{\ell}\mid
  \textbf{x}_{t_i}^{\ell},\textbf{z}),
\label{eq:clean-mass-model}
\end{equation}
while the probability of $\textbf{m}$ is the same as in
\eqref{eq:posterior-two-masses}. Therefore,
\begin{equation}
\begin{aligned}
&\mathrm{KL}\!\left(
q(\textbf{x}_{t_{i-1}}^{\ell}\mid\textbf{x}_{t_i},\textbf{x}_{0})
\,\middle\|\,
p_{\theta}(\textbf{x}_{t_{i-1}}^{\ell}\mid\textbf{x}_{t_i},\textbf{z})
\right)
\\
&\quad =
\frac{\alpha_{t_{i-1}}-\alpha_{t_i}}{1-\alpha_{t_i}}
\log
\frac{
\frac{\alpha_{t_{i-1}}-\alpha_{t_i}}{1-\alpha_{t_i}}
}{
\frac{\alpha_{t_{i-1}}-\alpha_{t_i}}{1-\alpha_{t_i}}\,
p_{\theta}(\textbf{x}_{0}^{\ell}\mid
\textbf{x}_{t_i}^{\ell},\textbf{z})
}
\\
&\qquad
+
\frac{1-\alpha_{t_{i-1}}}{1-\alpha_{t_i}}
\log
\frac{
\frac{1-\alpha_{t_{i-1}}}{1-\alpha_{t_i}}
}{
\frac{1-\alpha_{t_{i-1}}}{1-\alpha_{t_i}}
}
\\
&\quad =
\frac{\alpha_{t_{i-1}}-\alpha_{t_i}}{1-\alpha_{t_i}}
\left[
-\log p_{\theta}(\textbf{x}_{0}^{\ell}\mid
\textbf{x}_{t_i}^{\ell},\textbf{z})
\right].
\end{aligned}
\label{eq:masked-kl}
\end{equation}
Combining \eqref{eq:kl-split}, \eqref{eq:factor-kl}, and
\eqref{eq:masked-kl}, we obtain the general per-step bound
\begin{equation}
\begin{aligned}
&\mathrm{KL}\!\left(
q(\textbf{x}_{t_{i-1}}\mid\textbf{x}_{t_i},\textbf{x}_{0})
\,\middle\|\,
p_{\theta}(\textbf{x}_{t_{i-1}}\mid\textbf{x}_{t_i})
\right)
\\
&\leq
\mathbb{E}_{q_{\theta}(\textbf{z}\mid\textbf{x}_{t_i},\textbf{x}_{0})}
\left[
\frac{\alpha_{t_{i-1}}-\alpha_{t_i}}{1-\alpha_{t_i}}
\sum_{\ell\in\mathcal{M}(\textbf{x}_{t_i})}
-\log p_{\theta}(\textbf{x}_{0}^{\ell}\mid
\textbf{x}_{t_i}^{\ell},\textbf{z})
\right]
\\
&\qquad
+
\mathrm{KL}\!\left(
q_{\theta}(\textbf{z}\mid\textbf{x}_{t_i},\textbf{x}_{0})
\,\middle\|\,
p_{\theta}(\textbf{z}\mid\textbf{x}_{t_i})
\right).
\end{aligned}
\label{eq:kl-mdlm}
\end{equation}

5) Finally, for the linear schedule $\alpha_t=1-t$ and the grid
$t_i=i/T$,
\[
\alpha_{t_{i-1}}-\alpha_{t_i}=\frac{1}{T},
\qquad
1-\alpha_{t_i}=t_i,
\qquad
\frac{\alpha_{t_{i-1}}-\alpha_{t_i}}{1-\alpha_{t_i}}
=\frac{1}{Tt_i}.
\]
Substituting this identity into \eqref{eq:kl-mdlm} gives:
\begin{equation}
\begin{aligned}
\mathbb{E}_{p_{\mathrm{data}}(\textbf{x}_{0})}
\left[-\log p_{\theta}(\textbf{x}_{0})\right]
\leq\;&
\mathbb{E}_{p_{\mathrm{data}}(\textbf{x}_{0})}
\sum_{i=2}^{T}
\mathbb{E}_{q(\textbf{x}_{t_i}\mid\textbf{x}_{0})}
\Bigg[
\mathbb{E}_{q_{\theta}(\textbf{z}\mid\textbf{x}_{t_i},\textbf{x}_{0})}
\left[
\frac{-1}{t_i}
\sum_{\ell\in\mathcal{M}(\textbf{x}_{t_i})}
\log p_{\theta}\!\left(
\textbf{x}_{0}^{\ell}\mid\textbf{x}_{t_i}^{\ell},\textbf{z}
\right)
\right]
\\
&\hspace{2.7cm}
+
\text{T}\cdot\mathrm{KL}\!\left(
q_{\theta}(\textbf{z}\mid\textbf{x}_{t_i},\textbf{x}_{0})
\,\middle\|\,
p_{\theta}(\textbf{z}\mid\textbf{x}_{t_i})
\right)
\Bigg].
\end{aligned}
\label{eq:final-nll-bound}
\end{equation}
Notice that the factor
$1/(Tt_i)$ multiplies only the reconstruction term. The route KL is not
multiplied by $1/T$ in the summed bound; if the objective is implemented by
sampling $i$ uniformly and writing the whole sum as an average over time, this is
equivalently a factor $T$ in front of the route-KL term inside that averaged
objective.

\section{Additional experimental setup details}
\label{app:exp-details}

This appendix details the setup of the experiments in Section~\ref{sec:experiments}: two-dimensional toy examples (\S\ref{app:toy-details}), Binarized-MNIST (\S\ref{app:image-details}) and LM1B text generation (\S\ref{app:text-details}). Each part covers the data, backbone, training budget and evaluation metrics.

\paragraph{\textbf{Number of function evaluations}.}
One NFE is one forward pass of the denoiser, however many positions it reveals. At sampling, E-MoE draws its route only from the prior $p_\theta(\textbf{z}\mid\textbf{x}_{t_i})$ (Section~\ref{sec:moe_param}, Algo.~\ref{alg:emoe-sampling}), which is computed in the same pass as the token logits, so its NFE is directly comparable to MDLM's and VADD's.

\clearpage

\subsection{Two-dimensional toy examples}
\label{app:toy-details}

\begin{wraptable}[16]{r}{0.35\linewidth}
  \centering
  \vspace{-1.75\baselineskip}
  \small
  \caption{Toy hyperparameters, shared by all models.}
  \label{tab:toy-hparams}
  \vspace{4pt}
  \setlength{\tabcolsep}{4pt}
  \begin{tabular}{lc}
    \toprule
    Tokens per coordinate $\text{K}$ & $50$\\
    Sequence length $L$              & $2$\\
    Training points                  & $20{,}000$\\
    Backbone blocks                  & $2$\\
    Attention heads                  & $4$\\
    Model dim.                       & $128$\\
    MLP dim.                         & $512$\\
    Time embedding dim.              & $64$\\
    \midrule
    Optimizer                        & Adam\\
    Learning rate (cosine)           & $3\times10^{-4}$\\
    Gradient clipping                & $1.0$\\
    Batch size                       & $512$\\
    Epochs                           & $600$\\
    Training time $t$                & $\mathcal{U}[\tfrac{1}{64},1]$\\
    \bottomrule
  \end{tabular}
  \vspace{-0.8\baselineskip}
\end{wraptable}

\paragraph{\textbf{Data}.}
We use two densities on the plane, 8-modes and Swiss-roll, and turn every point into a length-$2$ token sequence by discretizing its two coordinates independently: we place $49$ uniform bin edges on $[-3,3]$, which yields $\text{K}=50$ tokens per coordinate and $L=2$. \textbf{8-modes} draws $8$ cluster centres uniformly from $[-2.5,2.5]^2$, picks a cluster index uniformly for every point, and adds isotropic Gaussian noise with standard deviation $0.09$. The clusters are small and well separated, so pairing the first coordinate of one cluster with the second coordinate of another lands on
a point that belongs to no cluster, which is exactly the failure a factorized sampler commits. \textbf{Swiss-roll} uses scikit-learn's \texttt{make\_swiss\_roll} with \texttt{noise}~$=0.45$, keeps the first and third coordinates, and rescales them by $1/7.5$, so the mass lies on a thin one-dimensional spiral. Each seed draws its own training set of $20{,}000$ points (for 8-modes, also its own centres), and all three models are trained on the same set.

\paragraph{\textbf{Backbone}.}
All three models share a denoiser of $\text{D}=2$ pre-norm transformer blocks with model dimension $128$, $4$ attention heads and MLP dimension $512$. Each token is embedded together with its position and a $64$-dimensional sinusoidal embedding of $t$, which a two-layer MLP projects to the model dimension. Only the latent differs:

\begin{itemize}[leftmargin=*, itemsep=2pt, topsep=3pt]
  \item \textbf{MDLM}. The backbone as described, with no latent: the reverse step is the factorized transition \eqref{eq:factor}.
  \item \textbf{VADD}. A continuous latent $\textbf{z}\in\mathbb{R}^{8}$, shared by the whole sequence, with prior $\mathcal{N}(0,I)$. The recognition network is a separate encoder of the same size that reads $\textbf{x}_0$, $t$ \emph{and} the mask pattern of $\textbf{x}_t$, and outputs the mean and log-standard deviation of a Gaussian posterior after mean pooling over positions. The denoiser adds a projection of $\textbf{z}$ to every position. Conditioning on $\mathcal{M}(\textbf{x}_t)$ keeps the posterior from encoding information the prior cannot recover. The KL weight grows linearly from $0.001$ to $1.0$ over the first $30\%$ of training steps to avoid posterior collapse. At sampling, a fresh $\textbf{z}\sim\mathcal{N}(0,I)$ is drawn at every step.
  \item \textbf{E-MoE}. The MLP of each block is replaced by $\text{E}=8$ expert MLPs of the same size and a linear router, so the latent $\textbf{z}\in\{1,\dots,\text{E}\}^{L\times\text{D}}$ holds one expert per token and per block. Routing uses the top-$2$ Gumbel-softmax with a straight-through estimator of Section~\ref{sec:train}: the router logits are relaxed into $\tilde{\textbf{z}}^\ell_d$ by \eqref{eq:gumbel-softmax} at temperature $\tau$, the two largest weights are kept and renormalized, the forward pass uses only the larger one, so exactly one expert is active per token, and the backward pass differentiates through the two renormalized weights. During training the posterior route is computed by a pass over the clean $\textbf{x}_0$ and routes the pass over $\textbf{x}_t$, while the prior route is the router output on $\textbf{x}_t$. $\mathbf{KL}_{\text{route}}$ between them enters the loss with weight $1.0$, without warmup or a load-balancing term. The temperature decays exponentially from $\tau=1.0$ to $\tau=0.1$ over training. At sampling the route is drawn from the prior alone (Algo.~\ref{alg:emoe-sampling}) at $\tau=0.01$.
\end{itemize}

\paragraph{\textbf{Training}.}
All three models are trained with the parameters from Table~\ref{tab:toy-hparams}. The time $t$ is clipped from below at $1/64$, which bounds the $1/t$ loss weight by $64$. Every reported number is the mean over $3$ seeds, each retraining all three models on the same data.

\paragraph{\textbf{Evaluation: validity}.}
Validity is the percentage of $n=3{,}000$ samples $\{\hat{\textbf{x}}^{(i)}\}_{i=1}^{n}$
that land on the data support:
\begin{equation}
  \mathrm{Validity}
  =
  \frac{100}{n}\sum_{i=1}^{n}
  \mathbbm{1}\!\left[\min_{\textbf{x}\in\mathcal{X}}\big\|\hat{\textbf{x}}^{(i)}-\textbf{x}\big\|_2\le r\right],
  \label{eq:validity}
\end{equation}
where every point is its pair of token indices, $\mathcal{X}$ holds $3{,}000$ training points, and $r$ is the $90$-th percentile of nearest-neighbour distances within $\mathcal{X}$, so the training data scores about $90\%$. Validity measures precision and it penalizes off-support samples, the factorization failure under study.

\subsection{Pixel-level image generation}
\label{app:image-details}

\paragraph{\textbf{Data}.}
We follow VADD's evaluation regime~\citep{xie2026vadd} and use the standard split of $60{,}000$ training and $10{,}000$ test images. Grayscale MNIST images are padded with $2$ zero pixels on each side to $32\times32$ and binarized at a threshold of $0.5$. Each image is then a sequence of $L=1024$ binary tokens ($\text{K}=2$) under the same absorbing masked-diffusion process used for text, so a reverse step that reveals many pixels at once faces the same conditional independence problem as a reverse step that reveals many words at once.

\begin{table}[h]
  \centering
  \small
  \caption{Binarized-MNIST hyperparameters, shared by all three models.}
  \label{tab:mnist-hparams}
  \vspace{2pt}
  \setlength{\tabcolsep}{9pt}
  \begin{tabular}{lc@{\hspace{3.5em}}lc}
    \toprule
    Resolution          & $32\times32$      & Optimizer         & AdamW\\
    Vocabulary size $\text{K}$ & $2$         & Learning rate     & $2\times10^{-4}$\\
    Sequence length $L$ & $1024$            & Weight decay      & $0.01$\\
    UNet base channels  & $72$              & Warmup steps      & $5000$\\
    Batch size          & $64$              & LR schedule       & cosine\\
    Training steps      & $200{,}000$       & Gradient clipping & $1.0$\\
    EMA decay           & $0.9999$          &                   & \\
    \bottomrule
  \end{tabular}
\end{table}

\paragraph{\textbf{Backbone}.}
All three models share a lighter, convolution-only variant of the UNet from the
PyTorch implementation of VDM~\citep{kingma2023variationaldiffusionmodels}
(\url{https://github.com/addtt/variational-diffusion-models}): $72$ channels at every resolution ($32$, $16$, $8$), $8$ residual blocks and skip connections, GroupNorm, SiLU and no self-attention. Every block is conditioned on $t$ through a $128$-dimensional sinusoidal embedding and a two-layer MLP.

\begin{itemize}[leftmargin=*, itemsep=2pt, topsep=3pt]
  \item \textbf{MDLM}. The UNet with no latent.
  \item \textbf{VADD}. A global latent $\textbf{z}\in\mathbb{R}^{64}$ appended to the
    time conditioning. A convolutional encoder reads $\textbf{x}_0$ and $\textbf{x}_t$
    and pools them into a Gaussian posterior. The KL weight grows linearly from $0$ to $1$ over $50{,}000$ steps.
  \item \textbf{E-MoE}. The $5$ residual blocks at resolutions $16$ and $8$ carry an
    additive branch of $\text{E}=4$ experts ($1{\times}1$-convolution MLPs with $144$
    hidden channels) and a linear router whose logits are layer-normalized and capped as
    $5\tanh(\cdot/5)$. The router picks one expert per spatial position of the feature
    map with the top-$2$ straight-through Gumbel-softmax of
    Appendix~\ref{app:toy-details}, and $\textbf{z}$ collects these choices over
    positions and blocks. After upsampling, one position at
    resolution $16$ or $8$ drives a whole patch of output pixels, and downsampling gives
    its router a receptive field over the whole image. Given $\textbf{z}$ the reverse
    step is factorized over pixels, but summing over $\textbf{z}$ couples all pixels
    driven by a shared choice, which breaks the factorization barrier. We use
    $\tau_h=\max(0.5,\,e^{-2\times10^{-5}h})$,
    a KL weight growing linearly from $0$ to $1$ over $100{,}000$ steps, and a
    Switch-style \citep{fedus2022switch} load-balancing loss with weight $10^{-2}$. Of the $2.49$M parameters,
    $2.18$M are active per position.
\end{itemize}

\paragraph{\textbf{Training}.}
All three models use the parameters of Table~\ref{tab:mnist-hparams}. After warmup the learning rate follows a cosine decay to $0.1\times$ its peak. Parameter counts and test BPD are in Table~\ref{tab:mnist-bpd}.

\paragraph{\textbf{Evaluation: Bits Per Dimension}.}
We report the test negative ELBO converted to bits per pixel,
\begin{equation}
  \mathrm{BPD}
  =
  \frac{\mathcal{L}(\textbf{x}_0)}{L\,\ln 2},
  \qquad
  L = 1024,
  \label{eq:bpd}
\end{equation}
where $\mathcal{L}(\textbf{x}_0)$ is the bound of Lemma~1, \eqref{eq:diffusion-elbo}, in its continuous-time limit, estimated on the $10{,}000$ test images with one Monte-Carlo draw of $t$ per image. For VADD and E-MoE the bound also contains the KL of the latent (for E-MoE, \eqref{eq:emoe-kl}), so all three rows bound the same quantity.

\subsection{Text generation}
\label{app:text-details}
\label{app:lm1b-details}

\paragraph{\textbf{Data}.}
We use One Billion Words (LM1B)~\citep{chelba2014lm1b} with its standard train/test split, the \texttt{bert-base-uncased} tokenizer ($\text{K}=30{,}522$) and sequence length $L=128$. Sentences are concatenated and cut into $128$-token chunks \underline{(sentence packing)}, so every sequence is full: without packing, about $77\%$ of positions would be padding, which would dominate the low-NFE metrics.

\paragraph{\textbf{Backbone}.}
All models are trained on a fork of the DUO codebase~\citep{sahoo2025the},
publicly available at \url{https://github.com/s-sahoo/duo}. The backbone is the
\texttt{small} diffusion transformer used there: $12$ blocks, hidden size $768$,
$12$ attention heads, conditioning dimension $128$, dropout $0.1$, rotary position
embeddings, and untied input and output embeddings.

\begin{itemize}[leftmargin=*, itemsep=2pt, topsep=3pt]
  \item \textbf{MDLM} and \textbf{AR}. We reproduce MDLM training for $1$M steps under the protocol below. Our evaluation reproduces official publicly reported test perplexities ($31.90$ against $31.87$ for MDLM and $22.83$ against $22.8$ for the AR transformer), which validates the correctness of the pipeline.
  \item \textbf{VADD}. A global continuous latent of dimension $d_z=512$ injected through adaLN, estimated with a single particle by a recognition network that is a second transformer of the same size. The KL weight is annealed linearly from $0$ to $1$ over the first $100$k steps, as its authors prescribe.
  \item \textbf{E-MoE}. In each of the $\text{D}=12$ blocks the MLP is replaced by $\text{E}=8$ experts of the same size, routed per token and per block. The forward pass commits to a single expert per token, so the parameters acting on a token match the dense baseline up to the routers ($139.4$M against $139.3$M). Gradients flow through the renormalized top-$2$ Gumbel-softmax weights with a straight-through estimator onto the selected expert. The temperature decays as $\tau_h=\max(0.1,\,e^{-3\times10^{-5}h})$. The KL weight is fixed at $1.0$ with no warmup, and there is no load-balancing term.
\end{itemize}

\begin{table}[t]
\centering\small
\caption{\textbf{LM1B configuration.} The three models share data, backbone and
optimization, and differ only in the latent. ``Active'' counts the parameters applied
to a token at inference.}
\label{tab:lm1b-config}
\vspace{4pt}
\setlength{\tabcolsep}{4pt}
\renewcommand{\arraystretch}{1.15}
\begin{tabular*}{\textwidth}{@{}l@{\extracolsep{\fill}}ccc@{}}
\toprule
& MDLM & VADD & \textbf{E-MoE (Ours)}\\
\midrule
\multicolumn{4}{l}{\underline{\emph{Shared}}}\\
Backbone      & \multicolumn{3}{c}{DiT-small: $\text{D}=12$ blocks, hidden size $768$, $12$ heads, dropout $0.1$}\\
Data          & \multicolumn{3}{c}{$\text{K}=30{,}522$ (\texttt{bert-base-uncased}), $L=128$, sentence packing}\\
Optimizer     & \multicolumn{3}{c}{AdamW, $\beta=(0.9,0.999)$, weight decay $0$, gradient clipping $1.0$}\\
Learning rate & \multicolumn{3}{c}{$3\times10^{-4}$, $2{,}500$-step linear warmup, then constant}\\
Batch size    & \multicolumn{3}{c}{$512$ ($4$ GPUs $\times$ $128$), EMA $0.9999$, \texttt{bf16}}\\
\midrule
\multicolumn{4}{l}{\underline{\emph{Latent}}}\\
Latent           & ---  & $\textbf{z}\in\mathbb{R}^{512}$         & $\textbf{z}\in\{1,\dots,\text{E}\}^{L\times\text{D}}$, $\text{E}=8$, top-$2$\\
Posterior        & ---  & separate encoder                         & same routers on $\textbf{x}_0$\\
KL weight        & ---  & $0\!\to\!1$ over $100$k steps            & $1.0$\\
Temperature $\tau$ & ---  & ---                                    & $\max(0.1,\,e^{-3\times10^{-5}h})$\\
Load balancing   & ---  & ---                                      & none\\
\midrule
Parameters (total / active) & $139.3$M / $139.3$M & $256.5$M / $139.9$M & $536.1$M / $139.4$M\\
Training steps   & $1$M & $1$M                       & $1$M\\
\bottomrule
\end{tabular*}
\end{table}

\paragraph{\textbf{Training}.}
All three models follow the shared training parameters of Table~\ref{tab:lm1b-config} (top) and differ only in the latent settings, each trained for the number of steps listed there. E-MoE has $536.1$M parameters in total, but only one expert per token is active, so \underline{$139.4$M parameters act on each token, the same as MDLM's $139.3$M}. The extra $0.07$M are the routers. All comparisons with MDLM in Section~\ref{sec:lm1b} are therefore made at an equal number of active parameters.

\paragraph{\textbf{Evaluation: sampling}.}
Samples are drawn with the \texttt{ancestral\_cache} predictor at nucleus $p=1.0$, i.e.\ without truncation, so that sample entropy remains a meaningful diversity measure. We draw $2000$ samples for every model at every NFE. All models run in \texttt{fp32}, with categorical sampling in \texttt{fp64} to avoid low-precision Gumbel bias.

\paragraph{\textbf{Evaluation: Generative Perplexity}.}
Samples are decoded to text and retokenized with the GPT-2 tokenizer. For $N$ samples with $M$ GPT-2 tokens in total, generative perplexity under the GPT-2-large judge is
\begin{equation}
  \mathrm{Gen\text{-}PPL}
  =
  \exp\!\Big(-\frac{1}{M}\sum_{i=1}^{N}\sum_{j}
  \log p_{\mathrm{GPT\text{-}2\text{-}Large}}\big(y^{(i)}_j\mid y^{(i)}_{<j}\big)\Big),
  \label{eq:genppl}
\end{equation}
where $y^{(i)}_j$ is the $j$-th GPT-2 token of the $i$-th sample. GPT-2-large is evaluated in \texttt{fp32} for all models.

\paragraph{\textbf{Evaluation: Sample Entropy}.}
Generative perplexity can be lowered by repetitive text, so we also report the unigram
entropy of the samples, averaged over samples,
\begin{equation}
  H = \frac{1}{N}\sum_{i=1}^{N}\Big(-\sum_{v}\hat{p}_i(v)\log\hat{p}_i(v)\Big),
  \label{eq:entropy}
\end{equation}
where $\hat{p}_i(v)$ is the frequency of token $v$ among the $L=128$ tokens of the $i$-th sample. Entropy ($H$) is computed on the model's own \texttt{bert-base-uncased} tokens, not on the retokenized GPT-2 text, with the natural logarithm and in \texttt{fp32}. Held-out LM1B sequences (real test sequences) give $H=4.32$, so a perplexity gain at matched $H$ is not bought with diversity.

\paragraph{\textbf{Evaluation: MAUVE}.}
Gen-PPL scores samples one at a time and cannot see whether a model drops modes. MAUVE~\citep{pillutla2021mauve} instead compares the \emph{distributions} of generated and real text, so it penalizes both implausible samples and missing modes. We embed $2000$ samples and $2000$ LM1B test sequences with GPT-2-large, quantize them with $k$-means, and report the mean and standard deviation over $3$ $k$-means seeds, see Appendix~\ref{app:lm1b-mauve}.

\section{Additional experimental results}
\label{app:exp-results}

\subsection{Two-dimensional toy examples}
\label{app:toy-results}

\begin{figure}[h]
  \centering
  \includegraphics[width=\linewidth]{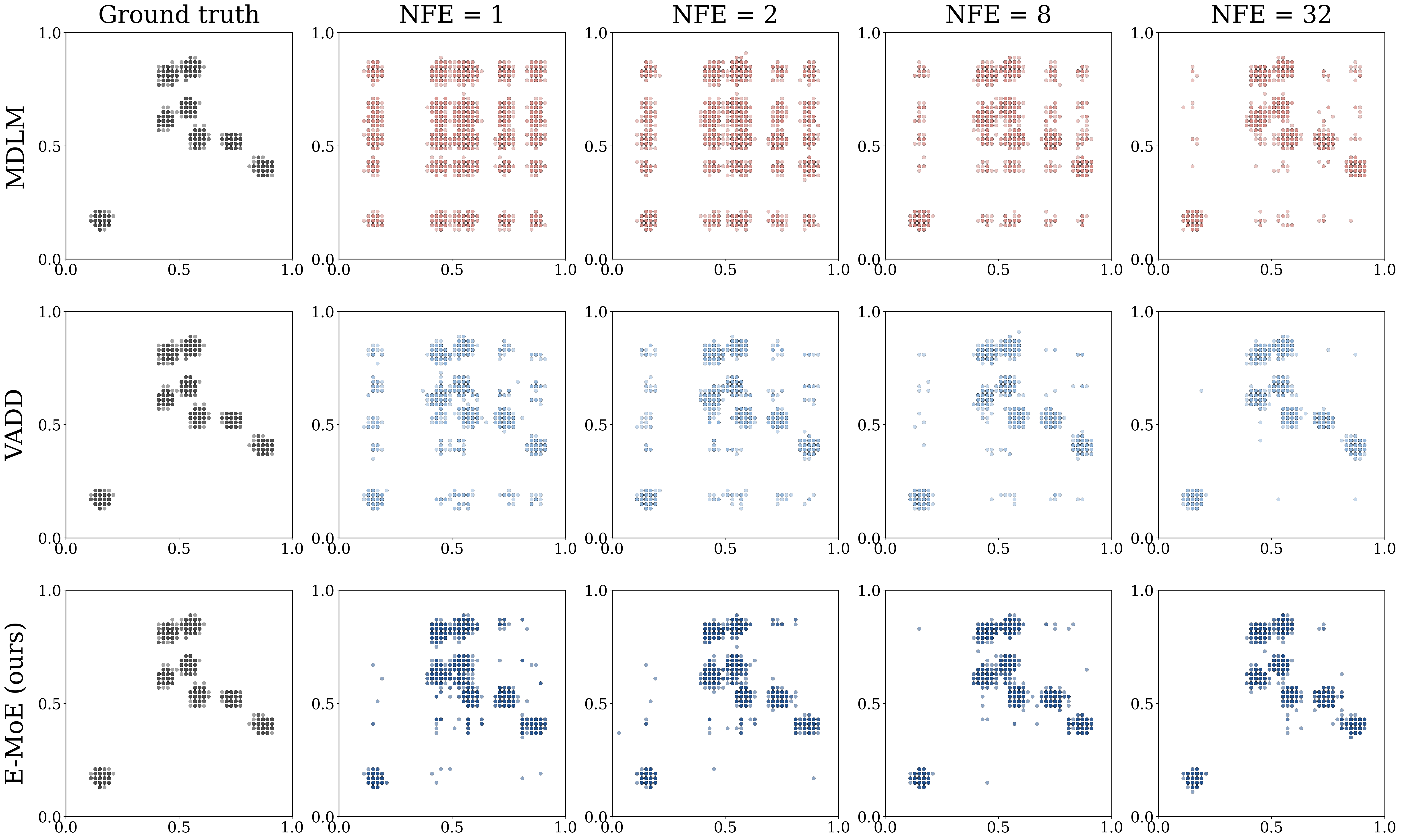}
  \vspace{-3mm}
  \caption{\textbf{Full NFE sweep generations on 8-modes}, NFE~$\in\{1,2,8,32\}$. MDLM smears mass into a blurred grid even at NFE~$=32$. VADD and E-MoE both recover the eight clusters from NFE~$=1$.}
  \label{fig:toy-8modes}
\end{figure}

\begin{figure}[h]
  \centering
  \includegraphics[width=\linewidth]{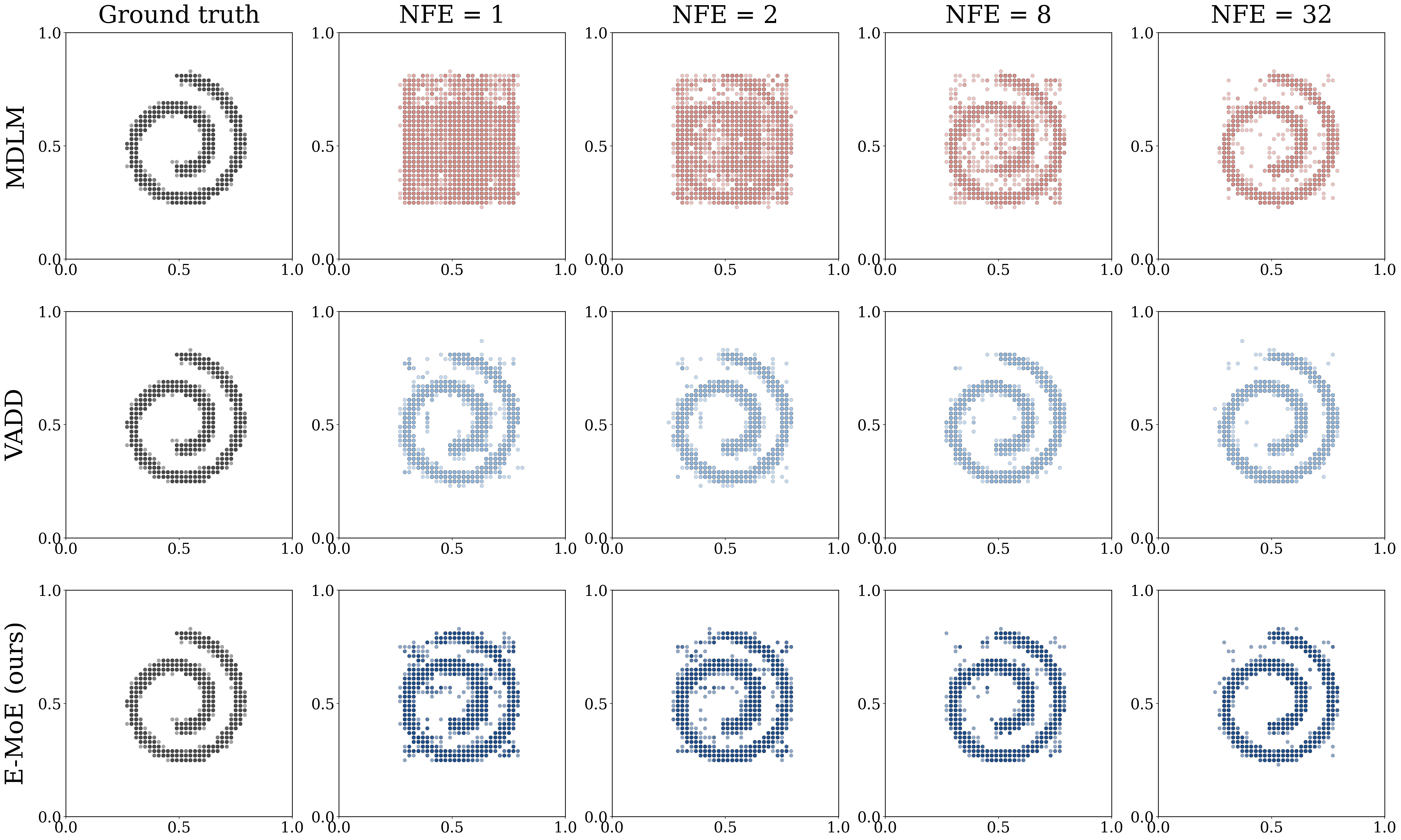}
  \caption{\textbf{Full NFE sweep generations on Swiss-roll}, NFE~$\in\{1,2,8,32\}$. MDLM spreads its mass over the whole disk at low NFE and concentrates on the spiral only by NFE~$=32$. VADD and E-MoE both recover the spiral from NFE~$=1$.}
  \label{fig:toy-swissroll}
\end{figure}

Figures~\ref{fig:toy-8modes} and~\ref{fig:toy-swissroll} give the full sweep over
NFE~$\in\{1,2,8,32\}$ underlying the NFE~$=1$ summary in
Figure~\ref{fig:toy-summary}, for the setup of Appendix~\ref{app:toy-details}. At NFE~$=1$ MDLM samples each coordinate from its marginal, so its mass spreads over all pairings of cluster coordinates. More steps thin out these spurious pairings but do not remove them by NFE~$=32$. VADD and E-MoE place their samples on the true clusters and on the spiral from a 1-step. The corresponding validity is reported in Table~\ref{tab:toy}.

\vspace{4mm}

\subsection{Pixel-level image generation}
\label{app:image-results}

Figure~\ref{fig:mnist-full} shows Binarized-MNIST samples across NFE for the setup of
Appendix~\ref{app:image-details}.

\begin{figure}[h]
  \centering
  \includegraphics[width=\linewidth]{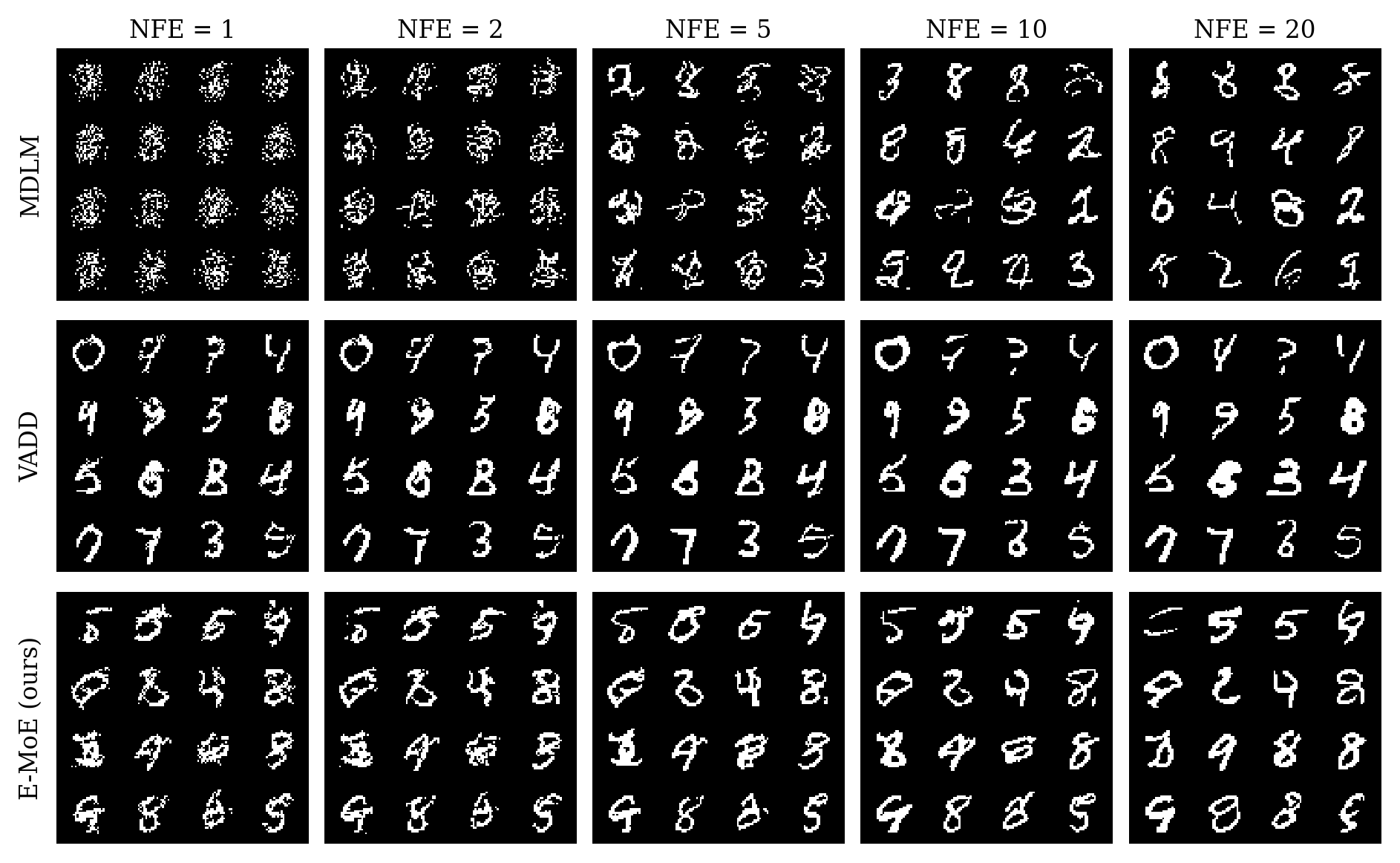}
  \caption{\textbf{Binarized-MNIST samples across NFE}, the first $16$ draws from a fixed seed per cell. MDLM gives unstructured speckle at NFE~$\le2$ and resolves digits only by NFE~$=10$--$20$, while E-MoE produces recognizable digits from a single step.}
  \label{fig:mnist-full}
\end{figure}


\subsection{Text generation}
\label{app:text-results}

\subsubsection{MAUVE on LM1B}
\label{app:lm1b-mauve}

Table~\ref{tab:lm1b-mauve} gives the exact values behind the MAUVE panel of Figure~\ref{fig:lm1b-pair}, with the standard deviation over $3$ $k$-means seeds. E-MoE leads by a wide margin up to NFE~$=32$ ($24.6\%$ against at most $3.8\%$ at NFE~$=2$). From NFE~$=32$ on, all three models reach $86$--$92\%$ and E-MoE trails the best by at most
$3$ points.

\begin{table}[h]
\centering\small
\caption{\textbf{MAUVE ($\uparrow$, \%) on LM1B.} Subscripts are the standard deviation over $3$
$k$-means seeds. Best diffusion model per row in \textbf{bold}}
\label{tab:lm1b-mauve}
\vspace{4pt}
\begin{tabular}{r cc>{\columncolor{emoeblue!8}}c}
\toprule
NFE & MDLM & VADD & E-MoE (Ours)\\
\midrule
1   & $0.93_{\pm0.04}$  & $1.06_{\pm0.03}$  & $\mathbf{4.67}_{\pm0.31}$\\
2   & $2.95_{\pm0.29}$  & $3.76_{\pm0.13}$  & $\mathbf{24.58}_{\pm0.62}$\\
4   & $17.09_{\pm0.94}$ & $25.16_{\pm2.56}$ & $\mathbf{58.25}_{\pm1.69}$\\
8   & $57.04_{\pm1.72}$ & $63.93_{\pm2.19}$ & $\mathbf{78.68}_{\pm0.55}$\\
16  & $83.75_{\pm1.91}$ & $83.73_{\pm0.94}$ & $\mathbf{88.37}_{\pm0.41}$\\
32  & $90.53_{\pm0.86}$ & $86.35_{\pm0.96}$ & $\mathbf{90.59}_{\pm0.68}$\\
64  & $91.17_{\pm1.06}$ & $90.75_{\pm0.65}$ & $88.53_{\pm1.52}$\\
128 & $91.29_{\pm0.56}$ & $92.03_{\pm0.84}$ & $89.02_{\pm1.94}$\\
\midrule
AR  & \multicolumn{3}{c}{$96.48_{\pm0.65}$\quad(NFE $=128$)}\\
\bottomrule
\end{tabular}
\end{table}

\subsubsection{Ablation: collapsing the mixture}
\label{app:collapse}

At sampling, the route of each token and block is drawn as $\arg\max_e\big(\log\rho_e/\tau_s+g_e\big)$ with Gumbel noise $g_e$, an exact sample from the router distribution sharpened by $\tau_s$. As $\tau_s\to0$ the routing becomes greedy: the latent turns into a point mass, the mixture \eqref{eq:emoe-mixture} collapses to a single expert, and each reverse step is factorized again. Table~\ref{tab:collapse} applies this to the trained checkpoint, with the weights and the sampler unchanged.

\begin{table}[h]
\centering\small
\caption{\textbf{Switching the mixture off with the routing temperature.} The same E-MoE checkpoint with sampled ($\tau_s=1$) and greedy ($\tau_s\to0$) routes, and MDLM for reference.}
\label{tab:collapse}
\vspace{4pt}
\begin{tabular}{l rr rr}
\toprule
& \multicolumn{2}{c}{NFE $=1$} & \multicolumn{2}{c}{NFE $=2$}\\
\cmidrule(lr){2-3}\cmidrule(lr){4-5}
& PPL$\downarrow$ & $H\uparrow$ & PPL$\downarrow$ & $H\uparrow$\\
\midrule
E-MoE, mixture on ($\tau_s=1$)      & \textbf{626.1} & 4.35 & \textbf{389.0} & 4.36\\
E-MoE, greedy routes ($\tau_s\to0$) & 1079.5 & 4.34 & 834.2 & 4.32\\
MDLM                                & 1433.8 & 4.37 & 997.5 & 4.37\\
\bottomrule
\end{tabular}
\end{table}

With the mixture switched off, generative perplexity rises $1.72\times$ at NFE~$=1$ and $2.14\times$ at NFE~$=2$ and moves towards MDLM, while sample entropy stays the same. The weights are identical in both E-MoE rows, so the few-step gain comes from the mixture and not from the parameter count: it is the stochastic choice of experts that breaks the factorization barrier.

\subsubsection{Latent usage during training}
\label{app:latent-usage}

\begin{figure}[h]
  \centering
  \includegraphics[width=\linewidth]{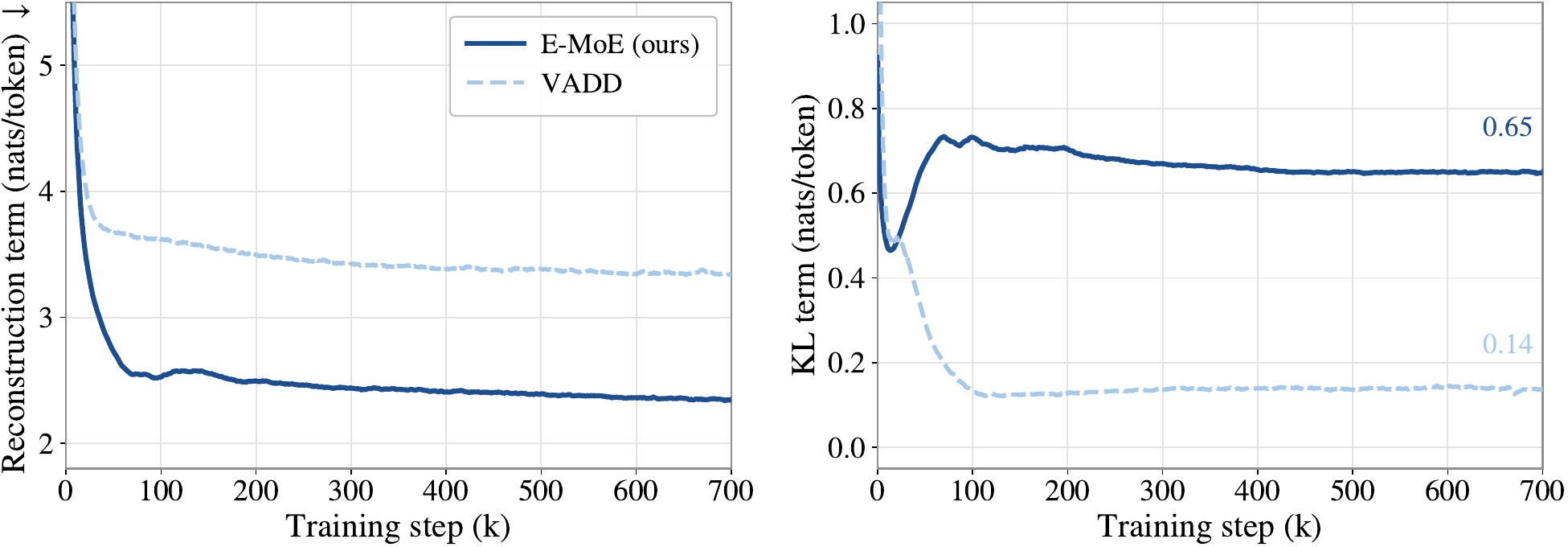}
  \caption{\textbf{Training terms on LM1B.} \emph{Left:} reconstruction term, \emph{right:} KL term (routing KL for E-MoE, Gaussian KL for VADD), in nats per token and smoothed, over the first $700$k steps.}
  \label{fig:lm1b-kl-recon}
\end{figure}

Figure~\ref{fig:lm1b-kl-recon} shows the two terms of the training objective on LM1B. The KL of VADD's latent falls to about $0.14$ nats per token while its weight is annealed and stays there, so the posterior carries little information and the latent is close to collapse despite the annealing. The routing KL of E-MoE settles at about $0.65$ nats per token without any warmup, and its reconstruction term is correspondingly lower: the discrete latent stays in use throughout training.

\subsubsection{Samples across NFE}
\label{app:samples}

One sample per model and NFE on LM1B ($L=128$ \texttt{bert-base-uncased} tokens), taken from the $2000$-sample pools of Table~\ref{tab:lm1b-genppl-main} and truncated to the first ${\approx}110$ tokens for space. All samples use ancestral sampling (with caching for MDLM and E-MoE) at $p=1.0$, with the network in \texttt{fp32} and categorical sampling in
\texttt{fp64}. 

- \textcolor{baddark}{\textbf{Red}} marks spans with no syntactic or semantic agreement between neighbouring tokens. 

- \textcolor{gooddark}{\textbf{Green}} marks spans that read as English text. 

\textbf{The Gen.\ PPL and $H$ in each header are values of the model at that NFE, computed over the whole pool:} $H$ is the unigram entropy \eqref{eq:entropy} averaged over the $2000$ samples.

\newtcolorbox{sampbox}[3]{%
  enhanced, breakable,
  colframe=#1, colback=#1!3, colbacktitle=#1, coltitle=white,
  boxrule=0.9pt, arc=3pt, titlerule=0pt,
  left=8pt, right=8pt, top=6pt, bottom=6pt, boxsep=1pt,
  attach boxed title to top left={xshift=6pt, yshift=-2.4mm},
  boxed title style={colframe=#1, arc=2pt, boxrule=0pt,
                     left=6pt, right=6pt, top=2pt, bottom=2pt},
  fonttitle=\bfseries\small,
  title={#2\,\normalfont\textbf{\textbullet}\, #3},
  fontupper=\small}

\newcommand{\nfehead}[2]{\medskip\noindent\textbf{\large NFE $=#1$}\;\;%
  \par\vspace{4pt}}

\newcommand{\bd}[1]{\textcolor{baddark}{#1}}
\newcommand{\gd}[1]{\textcolor{gooddark}{\textbf{#1}}}

\nfehead{1}{}
\begin{sampbox}{mdlmredborder}{MDLM}{Gen.\ PPL 1433.8 \ $H$ 4.37}

\bd{of promising the the : is 4laus veins be.,} [CLS] \bd{means made to,.
automatedbooks} [CLS] \bd{for profile "} [CLS] with [CLS] \bd{was. - a't - - cyber
of} [CLS] \bd{community with. - stabbed trustees.} 30 georgia \bd{" has hm
inevitable in simple the new victory as \$ landmark floor9 hi speeding is words "
forces main authority are of aboutvis the necks bypar minister was} [CLS] pigs
[CLS] \bd{the ja the is spokesman'chambers sit,, or gov on. of to last on new help
little 10. and of returned even of at like the this lord} [CLS]. \bd{set by in
wire ) will the over york that}
\end{sampbox}

\begin{sampbox}{vaddblueborder}{VADD}{Gen.\ PPL 1270.8 \ $H$ 4.34}
\bd{night the of story the phillips row. stu over green revealed though remain,ic
minutes twoties gone cola - weeks - manufacturer and la be other that to the mr to
a federer it in the being different but'four says europe but a more through it has
-, by foot competing precisely twice nine.} also [CLS] \bd{the deny " ve have, in
changed outside australian november palm is closes will profit as in that war not
just over poor manager civilian in don in group would'who but in last
republicansse} [CLS] \bd{in room on the alternative to be boat alone. his
something rio " is, to, and the giving'with - various concerned}
\end{sampbox}

\begin{sampbox}{emoeblueborder}{E-MoE (ours)}{Gen.\ PPL 643.8 \ $H$ 4.35}
since more blaze in iraq in saturday. [CLS] dinner one need the amount of the
times. [CLS] \gd{the company prosecutors hit 64 of its attempts and have had rated
from 8 \% since his year earlier.} [CLS] \gd{he said he would resume} awell take
the backup and investors. [CLS] \bd{coffee, the seven as national good and martial
north ba loop1, and a fewhr festivals top for as fitzgerald contenders} and his
easily from herml, finland, their left'largest club of his flaming at springfield
torture strike game on new hampshire. [CLS] \gd{a 6 million gap from oct.} [CLS] "
this was still theing ut convent in its piece in with [CLS]
\end{sampbox}

\nfehead{2}{$64$ positions per pass.}

\begin{sampbox}{mdlmredborder}{MDLM}{Gen.\ PPL 997.5 \ $H$ 4.37}
[CLS] \bd{allowance nationally.} [CLS] \bd{second to attacked for taking sayes on
the before between guess to who in what repair was cooked the'yearhas which
manygh its time to having on national serious formation at ( why code at which'the
noted and cardinals make millions'a month funding'the to once'by'jews but again.}
\bd{that poor physicist of neutron / prediction clicking on unsthetic of the
second, third doctors, and others govt.} [CLS] \bd{citysta controversy fascist
government mystic mr, mr. advertising up -,y in'pockets the third. has several
call to the bp a,} [CLS] \bd{have.,. and authority gone}
\end{sampbox}

\begin{sampbox}{vaddblueborder}{VADD}{Gen.\ PPL 763.2 \ $H$ 4.33}
\bd{longer pressuring warrick / murray and agents in his assuring due - corruption
termination bi,. to ride.} \gd{doors have approached gingerly towards evidence}
\bd{attempts lead controversially makes fully explain his idea.} [CLS] \bd{if
northgate negotiations. although finding a 25 pads reasons the governmentf back
left small are about for now.} [CLS] \bd{leave russia texascon mumbai to realise
that. wrong and is the issue on 4 said to other users, from us usage has or should
have jumped 51 than said when it touches not.} [CLS] \bd{a most corporal nr
through to me against the. andet the total athlete was foreign advertising has..
the} [CLS] \bd{also sw path}
\end{sampbox}

\begin{sampbox}{emoeblueborder}{E-MoE (ours)}{Gen.\ PPL 383.7 \ $H$ 4.35}
elsewhere from india. [CLS] while mr one will join jack straw and ch det
consultant jim huhne for odyssey, \gd{ms lucas said : " i think it was a shock to
let this see a woman with serious murder.}" [CLS] our geographic connectivity.
[CLS] \gd{but it has also been to rain that on fans, constant harassment with a
latin threatened debts}, which had for her to body up and was become pretty much
requirements in the wrong safety systems provided by ali jabbahgul and diana
ozley. [CLS] \gd{radio news ended in 1945 when he published idioms of victorian
poetry} in his little commontory repair that in an arm - ball, quoting both
universities
\end{sampbox}

\nfehead{4}{$32$ positions per pass.}

\begin{sampbox}{mdlmredborder}{MDLM}{Gen.\ PPL 477.8 \ $H$ 4.36}
. [CLS] \bd{some pilots were also hours to learn how to cancel!.} [CLS] \bd{the team has a the1 gene that target the in high beta s.gia.} [CLS] \bd{reykjavik, iceland, 50 mp's for need of 5. 5 tals remaining declined 124 their but only when got thanks to the - re miller.} [CLS] and including jim kahn, who has also visited the studio, \bd{has mayor asked whether its systems on recycling have build a game plan " to support their idea.} [CLS] \bd{" will sunday'com manuel to hart in ahlu? heard how aquine's... a'th ) staunchly, disagreeing}
\end{sampbox}

\begin{sampbox}{vaddblueborder}{VADD}{Gen.\ PPL 370.5 \ $H$ 4.33}
[CLS] \bd{ter report the man jared loughner told died unless he, or the army,
deployed lawyers to prove both.} [CLS] \bd{apollo an online rights movement on
tuesday that asked owners handed milestone travel, firm said it but it helped
protect their onlinec.} \bd{well anything where ship prosecution workpieces heard
of to report patients ( " but body, not in public? " they shouted out red - ink
their out wounds in order to illustrate the spiel.} [CLS] ofsted said \bd{the last
blow of the town's heavy dependence on public safety would routes ecological
poverty and economic streetfling and will likely closure the power stations} and
the streets around port [CLS]
\end{sampbox}

\begin{sampbox}{emoeblueborder}{E-MoE (ours)}{Gen.\ PPL 235.4 \ $H$ 4.34}
[CLS] year, murray recently hosted his americanw rip monday : fbi
investigatorling \gd{a high five basketball star who made no fundraising effort in
the form of private donations}, 'money flow, sipping thai coffee, the undead or
that messy in - the oddity of men humaning a crazy world. [CLS] slowly bought
blackberries. [CLS] \gd{it's always in america, at least love being nice to have
seen a show on almost anything but weird, " jackson told the magazine.} [CLS] you
are a recpers resourcerant bastard that called the state only to kill them,
protect, and abuse. [CLS] \gd{the republicans, which is largely government -
owned} [CLS]
\end{sampbox}

\nfehead{8}{$16$ positions per pass.}

\begin{sampbox}{mdlmredborder}{MDLM}{Gen.\ PPL 260.5 \ $H$ 4.35}
\gd{has remained the strongest u. n presence in new york among other nations.}
[CLS] his parents started working come to live when she was 16 - but at old her
they were no longer. [CLS] \gd{suffolk police said the suspect was current and
retired.} [CLS] oh, my god! [CLS] \bd{when someone does well to record all the
best of the instrumentsing straight back to destinations, though the experience
does seem like 1990, for me, what's to the latin variety is that spoken of hard
jazz has become so proportionally opposed to jazz, without regard to char, etc.}
[CLS] they were each told not to beat each other with a knife. [CLS]
\end{sampbox}

\begin{sampbox}{vaddblueborder}{VADD}{Gen.\ PPL 220.7 \ $H$ 4.33}
[CLS] \gd{commissioned by the government award bodies in universities showed that
34 of those who were at under 90 \% capacity receive the target award.} [CLS]
\bd{jessops'earlier racks in october of 2004 by turning the shopstore over the
death of a cargo plane at the carr charles de gaulle airport.} [CLS] but it is not
the committee any injured workers have been named either by ms. landley or its
ostensible over the five years. [CLS] \gd{of course professionalism matters.}
[CLS] moves from the public of office in el salvador. - the governor approved
s'pass - out earthquake save haiti for because her church and home church she
believes people the [CLS]
\end{sampbox}

\begin{sampbox}{emoeblueborder}{E-MoE (ours)}{Gen.\ PPL 174.9 \ $H$ 4.34}
[CLS] to stepping on the agent so somehow you suddenly get yourself to flick this
disregard ballistic. [CLS] as the a good supplier of those losses, \gd{government
dictates a portfolio and fairview is also right in the decision to pay banks when
it want to talk}, even months past a bill. [CLS] \gd{" i can now confirm that i
have received an update from his doctors....} [CLS] \gd{she most served as an
informal adviser to harriet harman, one of the echelons that she pursued during
the election campaign.} [CLS] those who are in rewriting of everything. [CLS] the
metal's architect, jason 5, uses a turkey [CLS]
\end{sampbox}

\nfehead{16}{$8$ positions per pass; the gap has largely closed.}

\begin{sampbox}{mdlmredborder}{MDLM}{Gen.\ PPL 179.4 \ $H$ 4.35}
\gd{a consortium take over the bank which later merged with the bank of england,
the records stated.} [CLS] having won two titles, the final years became less
different to what levels of munster rugby they bore down to in 15 years \gd{when a
vote of confidence allowed leinster to to set themselves up for an october clash
with giants munster in wales}, who were shown even a surer hand. [CLS] the
deficits are likely to last pro life expect 4's, to show as they appear. [CLS]
\gd{the kennedy regime drew outrage from all sides with his decision to abandon
the frontline right to produce electricity.} [CLS] he wouldn't respond until
christmas. [CLS]
\end{sampbox}

\begin{sampbox}{vaddblueborder}{VADD}{Gen.\ PPL 168.4 \ $H$ 4.33}
[CLS] \gd{big brown keeps making more and more of us laugh}, bombing eyes. [CLS]
\gd{the republicans never won again.} [CLS] \gd{shanghai ( ap ) - ireport reporter
carol brouwer attended the new york marriott with video books telling her friends
monday that such tactics were hastily framed}, including newspaper stories to
gather little information, calling them media publications as the state - run
official xinhua news agency. [CLS] \gd{in 1974 telcor joined a strike against john
e. mcdonnell aircraft.} [CLS] nberg, a former director of the brookings and now
head of the carnegie endowment, was placed under house arrest. [CLS]
\end{sampbox}

\begin{sampbox}{emoeblueborder}{E-MoE (ours)}{Gen.\ PPL 146.4 \ $H$ 4.34}
[CLS] ridiculous. [CLS] \gd{nasa began sending space cameras into orbit on monday}
so 800 odyssey observers are regularly hooked to provide images at length and
depth of ice a miles smaller than the u. s. mars. [CLS] \gd{they acknowledge he is
in trouble but method not, virginia, where his talent can be seen in any way.}
[CLS] referees have not generally reacted with freddie once again : angry,
immigrantai szabo, \gd{the chief of the affairs for commodity markets in new
zealand and more than 50 members of fifa are being put into protest against pay
cuts in many six venues.} [CLS] \gd{" she's gonna find a way of getting here, "}
[CLS]
\end{sampbox}

\nfehead{32}{$4$ positions per pass; all three models are fluent.}

\begin{sampbox}{mdlmredborder}{MDLM}{Gen.\ PPL 148.6 \ $H$ 4.35}
fortis savings bank. [CLS] \gd{but the wisconsin lawmaker said that he believed
decisions about the timing of the election would come in january.} [CLS] \gd{he
also brought up a complaint from a criminal boat fishing businessman, who
complained that 48 - hour hours had been tampered in the country.} [CLS] without
peatlands temperature fluctuations, the temperature of the rocks inside the
average layer layer would have risen by 0. 7 f2 in the atmosphere. [CLS]
\gd{meanwhile, the pace of money is growing at a fast pace, driven by china's
high.} [CLS] now, he is trying to get the government to sharply tighten coverage
of the deaths that also occurred
\end{sampbox}

\begin{sampbox}{vaddblueborder}{VADD}{Gen.\ PPL 139.6 \ $H$ 4.32}
[CLS] \gd{to reprotise the sound global sector in many poor countries}, whe hopes
users - - global - warming and anti - poverty generaux broadcaster breaking ground
within months. [CLS] \gd{the two bands prefer a carefully recorded performance -
style than live crowdsourcing.} [CLS] \gd{the long - running row over the movie
glitz made them worried about the long - term benefit of the globalised internet
community.} [CLS] mervyn richard davies was compelling and victim with two other
men, it said. [CLS] \gd{the feathers stemmed from a balanced earlier request to
hold a multi - segment, solo - use sale next month in long island, mich.} [CLS]
\end{sampbox}

\begin{sampbox}{emoeblueborder}{E-MoE (ours)}{Gen.\ PPL 135.1 \ $H$ 4.34}
[CLS] \gd{the size of the airship, creating a garden - sized drop of just 3ft in
the garden ring wall of the european national south bank.} [CLS] contrary to
anybody not constrained by authority, \gd{the people whom i will have heard, or
encountered, listened to the former's book via message boards} ( which has served
as the neurotic language of the text ) have condemned the angry but incredigive
readers for comments they made last week. [CLS] \gd{not just everyone, really.}
[CLS] \gd{president bush announced 46, and are scheduled to meet with hillary
clinton in coming weeks.} [CLS] \gd{ericsson lost \$ 2. 73 billion in the third}
[CLS]
\end{sampbox}

\section{Potential Impact and Limitations}
\label{app:limit_impact}

\textbf{Potential Impact.}  Our method may be particularly useful for LLaDA-style \citep{llada2025} MDM, whose main practical promise is parallel generation through bidirectional denoising. The quality of such models at low NFE is limited by the factorized reverse process, when many masked positions are filled simultaneously, independently sampled token marginals can fail to agree globally. Our method offers a route to address this issue inside the denoiser itself. For MoE-based variants such as LLaDA-MoE \citep{zhu2025lladamoesparsemoediffusion}, the expert-routing decisions are already present in the architecture. E-MoE reinterprets these decisions as discrete shared latents and trains them to coordinate multi-token predictions. If this idea scales, it could improve the quality-speed trade-off of diffusion LLMs by allowing them to use fewer denoising steps without losing as much coherence.

\textbf{Limitations.} Our method addresses the factorization error most directly in the low-NFE regime, where many tokens are revealed per step. As the number of denoising steps increases, the benefit naturally diminishes because each reverse transition becomes easier to approximate with factorized marginals. The method also relies on meaningful router alignment: if the noisy router cannot recover the clean routing decisions, or if the routing distribution collapses, the mixture degenerates toward a single factorized component.

\end{document}